\documentclass[11pt,a4paper]{article}
\usepackage[utf8]{inputenc}
\usepackage[T1]{fontenc}
\usepackage{amsmath,amssymb,amsthm}
\usepackage[margin=25mm]{geometry}
\usepackage{hyperref}
\hypersetup{colorlinks=true,linkcolor=blue,urlcolor=blue,citecolor=blue,pdftitle={Toward a Theoretical Understanding of the Transformer Architecture}}
\allowdisplaybreaks
\title{Toward a Theoretical Understanding of Transformer Attention: Bayesian Estimation and Predictive Sufficiency}
\author{Fouad B. Chedid\footnote{AI HORIZON, Muscat, Oman. fbchedid@gmail.com}}
\date{23 November 2026}
\begin{document}
\maketitle

\begin{abstract}
This paper develops a theoretical account of Transformer attention by organizing existing probabilistic and information-theoretic results. It identifies the assumptions under which attention computes a posterior mean and the additional conditions required for accurate next-token prediction. Posterior-mean optimality under squared-error loss does not by itself establish predictive sufficiency; even a sufficient representation requires a prediction head that uses its information correctly. An excess-log-loss decomposition makes this distinction precise by separating representation information loss from prediction-head error. The analysis clarifies the conditions required for predictive guarantees. Determining when these conditions hold in trained models, and how architectural or training choices can support them, could guide the development of AI systems that preserve and use predictive information more effectively.

\end{abstract}\tableofcontents\newpage

\section{Introduction}
Transformer-based large language models have transformed artificial intelligence, yet their predictive success has outpaced our theoretical understanding of the representations and computations that support it. Their training algorithm is explicit: a prediction objective guides parameter updates through backpropagation and numerical optimization. Knowing this algorithm, however, does not fully explain which internal representations emerge, which regularities they capture, or how they support accurate predictions on new inputs. This paper addresses a narrower question: under what assumptions does attention compute a Bayesian posterior mean, and under what additional conditions do its outputs support accurate next-token prediction? By connecting existing probabilistic interpretations of attention with predictive sufficiency and an excess-log-loss decomposition, we distinguish the conditions governing attention’s representations from those governing the prediction head’s use of them. The analysis clarifies requirements for predictive guarantees. Identifying architectural and training choices that help satisfy these conditions would connect the theoretical analysis to practical improvements in how AI systems retain and use information for prediction.

\subsection{The problem of next-token prediction}

One of the fundamental tasks performed by modern large language models is predicting the next token in a sequence. Consider the incomplete sentence:
\begin{quote}The Earth revolves around the ...\end{quote}
An effective language model assigns a high probability to a continuation such as “sun” and much lower probabilities to unrelated words.

This apparently simple task requires considerable contextual and conceptual information. The model must recognize grammatical structure, semantic relationships, factual associations, and dependencies among the observed words. Formally, let
\[ x_1,x_2,\ldots,x_t \]
denote a sequence of tokens from a vocabulary \(\mathcal X\). The objective is to estimate
\[ P(x_{t+1}\mid x_1,\ldots,x_t). \]
The general problem is therefore one of conditional prediction: given the information contained in an observed sequence, determine the probability distribution of the next symbol.

This problem predates modern artificial intelligence. It has deep connections to probability theory, statistical inference, information theory, and algorithmic information theory. Two developments are especially relevant to our investigation: Solomonoff induction and the Transformer architecture.

\subsection{Solomonoff induction and universal prediction}

Ray Solomonoff introduced a mathematical theory of prediction based on two ideas~\cite{solomonoff1964}. First, an observed sequence may be explained by different computational mechanisms. Second, simpler computational explanations should receive greater prior probability than more complicated ones. Solomonoff induction combines these ideas with Bayesian inference. It assigns weights to programs capable of generating observations and updates their relative support as additional symbols are observed. In this framework, prediction is obtained by averaging over competing generating mechanisms rather than committing prematurely to a single explanation.

Solomonoff induction provides important theoretical guarantees for computable sources of data, but its universal form is incomputable. It therefore represents a mathematical ideal rather than a directly executable prediction algorithm.

\subsection{The Transformer architecture}

The Transformer, introduced by Vaswani et al. in 2017, takes a different approach~\cite{vaswani2017}. Rather than explicitly enumerating possible generating programs, it learns numerical representations of tokens and uses attention to combine contextual information. The original architecture was designed for sequence-to-sequence tasks, particularly machine translation. Decoder-only variants subsequently became central to autoregressive language modeling. The central attention operation consists of three learned vector representations:
\begin{itemize}
\item Query (Q)
\item Key (K)
\item Value (V)
\end{itemize}
Queries and keys determine how strongly different positions in the input sequence interact. Values contain information that is combined according to the weights assigned to source positions through these query–key comparisons. This mechanism enables a Transformer to construct representations that depend on the current context.

\subsection{Research motivation}

The computational formula for attention is straightforward:
\[ \operatorname{Attention}(Q,K,V) = \operatorname{softmax} \left( \frac{QK^\top}{\sqrt{d_k}} \right)V. \]
Here, \(Q\) and \(K\) store query and key vectors as rows, so the \((i,j)\) entry of \(QK^\top\) is the dot product \(q_i^\top k_j\), comparing query \(i\) with key \(j\). The quantity \(d_k\) is the dimension of each query and key vector; dividing by \(\sqrt{d_k}\) helps keep these scores on a consistent numerical scale as \(d_k\) changes.

Yet the mathematical simplicity of this expression contrasts with the sophisticated predictive behavior of large Transformer models. 
This motivates a deeper question: What theoretical principle explains why evaluating relationships between learned representations and combining their associated information is so effective for prediction?

Our investigation begins with the observation that attention resembles probabilistic hypothesis evaluation. A query specifies the information currently sought. Each key identifies a candidate contextual information source. Their compatibility score determines the relative support assigned to that source. The associated value supplies information that may be incorporated into the resulting contextual representation.

Through the lens of Bayesian statistics, this suggests interpreting attention as a form of posterior-weighted information aggregation. The main objective of this paper is to develop that interpretation mathematically.

In-context learning provides a related setting: examples or instructions change a model's predictions without changing its pretrained parameters. Zhang et al.~\cite{zhang2025} establish connections to Bayesian model averaging and learning guarantees under specified statistical assumptions. These results support investigating Bayesian contextual computation, but do not transfer those guarantees automatically to general Transformer language models.

\subsection{Research questions and scope}

We address four questions.

\begin{enumerate}
\item Under what conditions can query--key compatibility be interpreted as evidence for a contextual hypothesis?
\item Under what conditions do attention weights correspond to Bayesian posterior probabilities over latent variables?
\item When can the weighted combination of value vectors be identified with a posterior expectation?
\item Can these operations help explain Transformer contextual prediction, representation refinement, and generalization?
\end{enumerate}
The first three questions admit exact answers under suitable probabilistic assumptions. The fourth requires a broader theory of representation learning and prediction. We therefore treat it as an open research problem.

\subsection{Structure of the paper}

The rest of the paper is organized as follows. Sections 2–4 introduce Bayesian inference, Solomonoff induction, and Transformer attention. Section 5 distinguishes posterior support from predictive influence. Section 6 develops the direct conditional and generative latent-source models, interprets value aggregation as a conditional expectation or posterior mean, and examines estimation optimality and stability. Section 7 investigates when these representations preserve the information needed for next-token prediction, separating representation loss from prediction-head error and exploring connections to information compression and algorithmic prediction bounds. Section 8 considers the implications for Transformer contextual prediction. Section 9 presents limitations and open questions, including the gap between computational capability and learned prediction, and Section 10 concludes the paper.

\section{Background: Bayesian inference}

\subsection{Hypotheses and evidence}

Bayesian inference provides a mathematical framework for updating beliefs when new information becomes available. Suppose we observe the sequence
\[ 2,\ 4,\ 6,\ 8. \]
Several hypotheses might explain it. One hypothesis proposes that the numbers are consecutive positive even integers. Another proposes a more complicated mathematical rule that happens to generate the same first four values. Bayesian inference does not require choosing one explanation immediately. Instead, it assigns probabilities to the competing hypotheses and updates those probabilities as observations accumulate. Let \(H\) represent a hypothesis and \(D\) the observed data. Bayes' rule states
\[  P(H\mid D) = \frac{P(D\mid H)P(H)}{P(D)}.   \]
The components have the following interpretations:
\begin{itemize}
\item \(P(H)\) is the prior probability of the hypothesis.
\item \(P(D\mid H)\) is the likelihood of the observations under the hypothesis.
\item \(P(D)\) is the marginal probability of the observations.
\item \(P(H\mid D)\) is the posterior probability.

\end{itemize}
For a finite hypothesis class \(\mathcal H\),
\[ P(D) = \sum_{h\in\mathcal H}P(D\mid h)P(h).  \]
The posterior is therefore determined jointly by prior plausibility and explanatory agreement with the observations.

\subsection{Sequential Bayesian updating}

Suppose observations arrive sequentially:
\[ x_1,x_2,\ldots,x_t. \]
Write \(x_{<t}=(x_1,\ldots,x_{t-1})\) for the observations preceding position \(t\), and \(x_{1:t}=(x_1,\ldots,x_t)\) for the observations up to and including position \(t\). After observing \(x_t\), the posterior becomes
\[  P_t(h) = \frac{ P_{t-1}(h)P(x_t\mid h,x_{<t}) }{ \sum_{h'} P_{t-1}(h')P(x_t\mid h',x_{<t}) }.   \]
Every observation modifies the probability assigned to competing hypotheses, and the updated posterior becomes the prior for the next observation. For the complete sequence,
\[ P_t(h) = \frac{ P(h)P(x_{1:t}\mid h) }{ P(x_{1:t}) }.  \]
Using the probability chain rule,
\[ P(x_{1:t}\mid h) = \prod_{i=1}^{t} P(x_i\mid h,x_{<i}).  \]

Consequently,
\[ P_t(h) = \frac{ P(h)\prod_{i=1}^{t}P(x_i\mid h,x_{<i}) }{ P(x_{1:t}) }.   \]
s
Taking logarithms yields
\[ \log P_t(h) = \log P(h) + \sum_{i=1}^{t} \log P(x_i\mid h,x_{<i}) - \log P(x_{1:t}). \tag{1}\label{eq:paper-7} \]

Equation~\eqref{eq:paper-7} reveals that posterior inference incorporates accumulated evidence in the form of log-likelihood contributions.

\subsection{Bayesian model averaging}

Once posterior probabilities have been computed, prediction is obtained by averaging the conditional predictions of the hypotheses:
\[
P(x_{t+1}\mid x_{1:t})
=\sum_{h\in\mathcal H}P_t(h)P(x_{t+1}\mid h,x_{1:t}).
\]
This operation is known as Bayesian model averaging. Each hypothesis contributes its next-token prediction, weighted by its posterior probability. The resulting prediction therefore depends both on the support for each hypothesis and on what that hypothesis predicts.

Sections 2.2–2.3 highlight two mathematical operations fundamental to Bayesian prediction:
\begin{enumerate}
\item Bayesian conditioning updates the probabilities of competing hypotheses.
\item Posterior-weighted aggregation combines their conditional predictions.
\end{enumerate}

These operations will reappear in our probabilistic interpretation of attention.

\section{Background: Solomonoff induction}

\subsection{Why prior probabilities matter}

Bayesian inference requires a prior distribution over hypotheses. But how should such priors be assigned?

Solomonoff induction addresses this problem by representing explanations as computer programs and assigning greater prior weight to shorter descriptions. The underlying principle is closely related to Occam's razor: among explanations consistent with the data, simpler explanations should generally receive greater initial preference.

\subsection{Prefix Kolmogorov complexity}

Prefix Kolmogorov complexity measures the length of a shortest computational description~\cite{liVitanyi}. Let \(U_{\mathrm{pf}}\) be a fixed universal prefix-free Turing machine, whose halting programs are self-delimiting: no valid halting program is a proper prefix of another. For a finite binary string \(x\), define
\[
K(x)=\min_{p:U_{\mathrm{pf}}(p)=x}|p|. 
\]
Here, \(U_{\mathrm{pf}}(p)=x\) means that the machine outputs exactly \(x\) and halts, and \(|p|\) is the program length in bits.

Regular strings can have short descriptions. For example, \(n\) repetitions of \(01\) can be described by a repetition procedure and the number \(n\). Most binary strings of a given length, however, cannot be described substantially more briefly than by specifying their bits.

We also use this notion for integers and computable probability measures. The quantity \(K(n)\) is the complexity of a standard binary encoding of the integer \(n\), while \(K(\mu)\) is the length of a shortest self-delimiting description of a procedure computing the probability measure \(\mu\). Conditional complexity \(K(\mu\mid x)\) measures the shortest such description when \(x\) is supplied as additional input. A procedure computing \(\mu\) takes a finite string and a requested precision as inputs and returns an approximation to its probability with that precision; it need not output an entire infinite probability distribution and halt. 

The prefix-free machine used here is distinct from the universal monotone machine used below to define Solomonoff's semimeasure \(M\). We use separate symbols to distinguish the two constructions.

\subsection{The universal semimeasure}

Let \(U_{\mathrm{mon}}\) be a fixed universal monotone Turing machine. It reads binary input in one direction and produces output without revising symbols already written. Unlike the prefix-free machine used in the previous section, it need not halt after producing the observed finite string, and its possible inputs need not form a prefix-free set.

Write \(x\preceq U_{\mathrm{mon}}(p)\) to mean that the machine produces an output beginning with \(x\) after reading no more than the input bits in the finite string \(p\). For a fixed observed string \(x\), define the set of minimal sufficient input prefixes by
\[
\mathcal P_x=\left\{
\begin{array}{l|l}
p\in\{0,1\}^{*} &
\begin{gathered}
x\preceq U_{\mathrm{mon}}(p),\quad\text{and}\\
x\npreceq U_{\mathrm{mon}}(q)\text{ for every proper prefix }q\text{ of }p
\end{gathered}
\end{array}
\right\}.
\]
Here, minimal means that no shorter input prefix of the same string already suffices; it does not mean shortest among all descriptions of \(x\).

For each fixed \(x\), the set \(\mathcal P_x\) is prefix-free. If one member were a proper prefix of another, the longer member would not be minimal. However, these prefixes are not necessarily self-delimiting programs in the sense of the previous section: their minimality depends on the particular observed output \(x\), rather than on a globally prefix-free domain of halting programs.

Suppose the machine is supplied with an infinite stream of independent, uniformly random input bits. The probability that this stream begins with a specified prefix \(p\) is \(2^{-|p|}\). Because \(\mathcal P_x\) is prefix-free, the corresponding input events are disjoint, so their probabilities can be added without double-counting. Solomonoff's universal semimeasure is
\[  M(x)=\sum_{p\in\mathcal P_x}2^{-|p|}.   \]
Thus, \(M(x)\) is the probability that the output of \(U_{\mathrm{mon}}\) begins with \(x\) under fair random input. Each additional bit halves an individual prefix's weight, implementing a preference for shorter descriptions while retaining the contributions of all minimal sufficient prefixes. Producing \(x\) does not require producing any further output. After producing \(x\), the machine may halt, run indefinitely without further output, or continue producing symbols. For binary outputs this gives
\[
M(x0)+M(x1)\leq M(x).
\]
The difference is the probability of input streams that produce \(x\) but never produce a subsequent output bit. This possible loss of continuation mass makes \(M\) a semimeasure rather than, in general, a probability measure on infinite output sequences.

\subsection{Solomonoff sequential prediction}

The conditional continuation weight below follows from the definition of conditional probability, applied to events about the machine’s output. We use the term “weight” to emphasize that \(M\) is a semimeasure, so the weights assigned to possible next symbols may sum to less than one. Let \(E_x\) be the event that its output begins with \(x\), and let \(E_{xy}\) be the event that it begins with the concatenation \(xy\). Since \(E_{xy}\subseteq E_x\) (any output beginning with \(xy\) necessarily begins with \(x\)),
\[
P(E_{xy}\mid E_x)
=\frac{P(E_{xy})}{P(E_x)}
=\frac{M(xy)}{M(x)},
\]
provided \(M(x)>0\). Here, probabilities refer to independent, uniformly random input bits supplied to the machine. Thus, given an observed prefix \(x\), we define the conditional continuation weight of a finite string \(y\) by
\[
M(y\mid x)=\frac{M(xy)}{M(x)}.
\]

For the next binary symbol,
\[
M(0\mid x)+M(1\mid x)\leq1.
\]
The missing mass corresponds to the possibility of no subsequent output. 

Solomonoff induction combines the simplicity-weighted construction with Bayesian conditioning. As additional symbols are observed, conditioning changes the relative support assigned to compatible computational explanations. Prediction aggregates their contributions rather than committing to one explanation. 

The universal semimeasure \(M(x)\) is incomputable, a limitation closely related to the undecidability of the halting problem. Computing \(M(x)\) requires summing the weights of all minimal input prefixes that cause a universal monotone Turing machine to produce an output beginning with \(x\). Although simulation can reveal contributions as they occur, no general algorithm can determine whether every remaining computation will eventually produce such an output or continue indefinitely without doing so. Hence, all that we can do is to approximate \(M(x)\) from below by progressively discovering contributions, obtaining increasingly accurate lower bounds. Thus, \(M(x)\) is lower semicomputable, but not computable. Still, \(M(x)\) has strong theoretical prediction properties. In particular, for any computable probability measure \(\mu\), there is a constant \(c_\mu>0\), independent of \(x\), such that
\[
M(x)\geq c_\mu\mu(x). \tag{2}\label{eq:universal-dominance}
\]
This property is called universal dominance. It says that, for every computable probability measure \(\mu\), the universal semimeasure \(M\) assigns each finite string \(x\) at least a fixed positive fraction of its probability under \(\mu\). Its predictive significance can be expressed through bounds on cumulative excess log loss. This is how.

Logarithmic loss measures the error of a probabilistic prediction. After observing an outcome, this loss is the negative logarithm of the probability assigned to that outcome; assigning a smaller probability incurs a larger loss. Define the cumulative excess log loss over \(n\) observations by
\[
\begin{aligned}
L_n
&=\sum_{t=1}^{n}\log_2
\frac{\mu(X_t\mid X_{<t})}{M(X_t\mid X_{<t})}.
\end{aligned}
\]
Here, \(X_t\) is the random symbol at position \(t\), and \(X_{<t}=(X_1,\ldots,X_{t-1})\) is the preceding random sequence. Lowercase \(x_t\) denotes a particular observed value of \(X_t\). In language modeling, these symbols correspond to tokens from the model's vocabulary. We use the convention \(M(\epsilon)=1\) for the empty string \(\epsilon\). The quantity \(L_n\) compares the cumulative log loss of the semimeasure-based predictor with that of a predictor that already knows the true generating distribution \(\mu\). At step \(t\), subtracting the true predictor's loss from Solomonoff's loss gives
\[
\begin{aligned}
&-\log_2 M(X_t\mid X_{<t}) 
-\bigl[-\log_2\mu(X_t\mid X_{<t})\bigr]\\
&\qquad=\log_2\frac{\mu(X_t\mid X_{<t})}{M(X_t\mid X_{<t})}.
\end{aligned}
\]
Summing these differences over the observations gives \(L_n\). Siince a sum of logarithms is the logarithm of a product,
\[
L_n=\log_2
\frac{\prod_{t=1}^{n}\mu(X_t\mid X_{<t})}
{\prod_{t=1}^{n}M(X_t\mid X_{<t})}.
\]
The probability chain rule gives \(\prod_{t=1}^{n}\mu(X_t\mid X_{<t})=\mu(X_{1:n})\). For \(M\), the definition of its conditional continuation weights gives:
\[
\prod_{t=1}^{n}M(X_t\mid X_{<t})
=\prod_{t=1}^{n}\frac{M(X_{1:t})}{M(X_{<t})}
=\frac{M(X_{1:n})}{M(\epsilon)}
=M(X_{1:n}).
\]
Therefore,
\[
L_n=\log_2\frac{\mu(X_{1:n})}{M(X_{1:n})}
=-\log_2 M(X_{1:n})+\log_2\mu(X_{1:n}).
\]

The same true-to-predicted log-ratio comparison will reappear in our probabilistic interpretation of attention.

\paragraph{From universal dominance to a cumulative excess-loss bound.}
Universal dominance gives a bound for every finite sequence of positive probability under \(\mu\). Applying \(-\log_2\) to Equation~\eqref{eq:universal-dominance} reverses the inequality:
\[
 \quad\Longrightarrow\quad
-\log_2 M(x_{1:n})\leq-\log_2\mu(x_{1:n})-\log_2 c_\mu.
\]
Subtracting the true predictor’s sequence loss, \(-\log_2\mu(x_{1:n})\), from both sides and using the definition of \(L_n\) gives
\[
L_n\leq-\log_2 c_\mu.
\]
Because this holds for every sequence of positive probability under \(\mu\), it also holds after averaging over sequences generated by \(\mu\):
\[
\mathbb E_\mu[L_n]\leq-\log_2 c_\mu.
\]
Since this upper bound is independent of \(n\), the expected excess log loss per symbol tends to zero as the number of observations grows:
\[
0\leq\frac{\mathbb E_\mu[L_n]}{n}
\leq\frac{-\log_2 c_\mu}{n}
\xrightarrow[n\to\infty]{}0.
\]
This result says that as more observations arrive, Solomonoff’s predictor incurs, on average, less additional log loss per symbol compared with a predictor that knows the true generating process. This does not mean that every next symbol becomes predictable with certainty: the generating process may remain random. The result concerns the accuracy of the predicted probabilities, rather than whether a representation preserves all information needed for prediction.

\paragraph{Convergence to the true predictive probabilities.}
Solomonoff’s predictor has a strong theoretical property: for a computable data-generating probability measure, its next-symbol predictions converge to the true conditional probabilities with probability one. This result concerns learning the probabilities of outcomes, not predicting every outcome with certainty \cite{hutter2005}. It applies to the ideal, incomputable predictor and does not automatically extend to Transformers or other computable predictors.

\paragraph{Relevance to Transformer prediction.}

Solomonoff induction and Transformer language modeling share an objective:
\[ \text{Observed sequence} \longrightarrow \text{Next-symbol prediction}. \]

However, their computational methods differ fundamentally. Solomonoff induction averages over possible generating programs. A Transformer learns a finite family of vector transformations and combines contextual representations. There is no general mathematical equivalence between the two systems. Nevertheless, Solomonoff induction provides a useful conceptual starting point: prediction can be understood as combining explanations according to evidence. The remainder of the paper investigates under what assumptions Transformer attention can implement a tractable form of this general principle.

\section{Background: The Transformer architecture}

\subsection{Token embeddings}

A Transformer language model receives a sequence of tokens
\[
x_1,x_2,\ldots,x_t.
\]
Each token is mapped to a learned embedding in a \(d\)-dimensional space. Let the embedding of token \(x_i\) be denoted by
\[
e_i=E(x_i)\in\mathbb R^d.
\]

Positional information is also incorporated. In an additive positional-embedding formulation, the position \(i\) is represented by a vector
\[
p_i\in\mathbb R^d,
\]
which is computed using a fixed encoding. The initial input representation is then
\[
h_i^{(0)}=e_i+p_i. 
\]

Here, \(i\) identifies the token's position, \(e_i\) is its token embedding, and \(p_i\) represents its position. The superscript \((0)\) labels the initial representation, before any Transformer layers have processed it. More generally, \(h_i^{(\ell)}\) denotes the representation at position \(i\) after \(\ell\) Transformer layers.

\subsection{Query, Key, and Value}

For each token representation \(h_i\), the attention mechanism constructs three vectors through learned linear projections: the query \(q_i\), the key \(k_i\), and the value \(v_i\):
\[ q_i=W_Qh_i,  \]
\[ k_i=W_Kh_i,  \]
\[ v_i=W_Vh_i.  \]

The projection matrices \(W_Q\), \(W_K\), and \(W_V\) are learned during training. The input representation has dimension \(d\), whereas the queries and keys have dimension \(d_k\), and the values have dimension \(d_v\):
\[
h_i\in\mathbb R^d,\qquad q_i,k_i\in\mathbb R^{d_k},\qquad v_i\in\mathbb R^{d_v}.
\]
Using a column-vector convention, we have
\[
W_Q,W_K\in\mathbb R^{d_k\times d},\qquad W_V\in\mathbb R^{d_v\times d}.
\]
The subscript \(k\) stands for key. Queries and keys must have matching dimensions so that their dot product is defined; values need not have the same dimension. When \(d_k<d\), these query and key projections perform dimensionality reduction: they learn to combine the input components into fewer output components. For a sequence of length \(t\), computing all query--key dot products costs on the order of \(t^2d_k\), rather than \(t^2d\) for full-dimensional queries and keys. The score matrix still has \(t^2\) entries in full attention, so this reduction does not remove the quadratic dependence on sequence length. The sequence length \(t\) is bounded by the Transformer's context-window capacity.

In multi-head attention, \(d_k\) and \(d_v\) describe the query/key and value dimensions, respectively, within each head. A common choice with \(m\) heads is \(d_k=d_v=d/m\). For example, the 175-billion-parameter GPT-3 model~\cite{brown2020} has an embedding dimension of \(d=12{,}288\) and \(m=96\) attention heads per layer, giving \(d_k=d_v=128\): each head uses 128-dimensional query, key, and value vectors. Splitting a fixed total dimension among several heads allows different learned projections without multiplying the main attention computation by the number of heads, compared with a single full-dimensional head. With this choice, the total projection parameter count and the computation required for score calculation and value aggregation are approximately the same as for a single full-dimensional head.

The vectors \(q_i\), \(k_j\), and \(v_j\) serve distinct computational roles. For a query position \(i\) and an accessible source position \(j\), the query \(q_i\) represents the information sought at position \(i\). The key \(k_j\) is compared with \(q_i\) to determine the source's attention weight relative to other accessible sources. The value \(v_j\) supplies the information combined according to that weight. The query and source positions may coincide; that is, a token can attend to its own position.

\subsection{Query--Key compatibility}

For query position \(i\) and candidate source position \(j\), the compatibility score is
\[  s_{ij} = \frac{q_i^\top k_j}{\sqrt{d_k}}.   \]
The dot product adds together \(d_k\) products of query and key components:
\[
q_i^\top k_j=\sum_{r=1}^{d_k}q_{ir}k_{jr}.
\]
As the vectors get longer, these sums tend to fluctuate more widely. Assuming that the query and key components are standardized (having mean zero and variance one) and are mutually independent, the dot product has variance \(d_k\), so its standard deviation is \(\sqrt{d_k}\). Dividing by \(\sqrt{d_k}\) gives the scaled score a variance of one, keeping scores on a roughly consistent scale across different vector dimensions.

The resulting scores are passed through softmax:
\[  \alpha_{ij} = \frac{\exp(s_{ij})} {\sum_{r\in\mathcal A_i}\exp(s_{ir})},   \]
where \(\mathcal A_i\) is the set of positions accessible to the query at position \(i\); that is, the positions that token \(i\) is allowed to attend to. In causal self-attention, \(\mathcal A_i\) includes all earlier positions and position \(i\) itself:
\[
\mathcal A_i=\{1,\ldots,i\}.
\]
Later positions are excluded, preventing the model from using future tokens when predicting the next token. The softmax outputs are called attention weights. They are nonnegative and sum to one over the accessible source positions:
\[ \alpha_{ij}\geq0, \qquad \sum_{j\in\mathcal A_i}\alpha_{ij}=1.  \]

\subsection{Value aggregation}

For each query position \(i\), attention produces an output vector \(z_i\) by taking a weighted average of the value vectors at accessible source positions:
\[
z_i=\sum_{j\in\mathcal A_i}\alpha_{ij}v_j. \tag{3}\label{eq:paper-19} 
\]
Here, \(v_j\) supplies the information associated with source position \(j\). The attention weight \(\alpha_{ij}\) determines how much that source contributes for query position \(i\). By contrast, \(z_i\) is the attention output vector, obtained by combining the value vectors according to those weights. Thus, the attention weights specify the contributions, while the attention output contains the aggregated information. In matrix notation,
\[  Z=\operatorname{Attention}(Q,K,V)= \operatorname{softmax} \left( \frac{QK^\top}{\sqrt{d_k}}+M_{\mathrm{mask}} \right)V.  \tag{4}\label{eq:paper-20} \]

The \(i\)-th row of \(Z\) is \(z_i^\top\). Each row of \(Q\), \(K\), and \(V\) contains the corresponding query, key, or value vector, respectively, and softmax is applied separately to each row of the score matrix. The additive mask \(M_{\mathrm{mask}}\) specifies which source positions are allowed for each query.

A causal mask implements causal self-attention by setting
\[
(M_{\mathrm{mask}})_{ij}=
\begin{cases}
0, & j\leq i,\\
-\infty, & j>i.
\end{cases}
\]
Adding zero leaves an allowed score unchanged. Adding \(-\infty\) to a future-position score makes its softmax weight zero. Consequently, \(\alpha_{ij}=0\) whenever \(j>i\), and Equation~\eqref{eq:paper-20} reproduces the restricted sum in Equation~\eqref{eq:paper-19}.

\subsection{Multi-head attention}

Let \(H\in\mathbb R^{t\times d}\) contain the token representations as rows. Each of \(m\) attention heads uses its own learned projections:
\[
Q^{(r)}=H(W_Q^{(r)})^\top,\qquad
K^{(r)}=H(W_K^{(r)})^\top,\qquad
V^{(r)}=H(W_V^{(r)})^\top.
\]
The transposes reconcile this row representation with the column-vector convention in Section 4.2. Each head applies the attention operation and mask defined above:
\[ Z^{(r)}=\operatorname{Attention}(Q^{(r)},K^{(r)},V^{(r)}).  \]
The outputs are concatenated and projected back to the model dimension:
\[ \operatorname{MHA}(H)=\operatorname{Concat}(Z^{(1)},\ldots,Z^{(m)})W_O.  \]
Each head returns one \(d_v\)-dimensional output per token:
\[
Z^{(r)}\in\mathbb R^{t\times d_v}.
\]
Concatenation places the heads’ outputs side by side for each token, preserving the number of rows:
\[
\operatorname{Concat}(Z^{(1)},\ldots,Z^{(m)})
\in\mathbb R^{t\times(md_v)}.
\]
The learned output projection has dimensions
\[
W_O\in\mathbb R^{(md_v)\times d}.
\]
Therefore,
\[
\underbrace{\operatorname{MHA}(H)}_{t\times d}
=
\underbrace{\operatorname{Concat}(Z^{(1)},\ldots,Z^{(m)})}_{t\times(md_v)}
\underbrace{W_O}_{(md_v)\times d}.
\]
Multiplication by \(W_O\) mixes the heads’ contributions and returns each token’s representation to dimension \(d\).  

Multiple attention heads provide different learned projections through which the model can detect and combine contextual regularities. This resembles the use of multiple filters in a Convolutional Neural Network, although attention weights adapt to the current input, whereas a standard convolution filter applies fixed learned coefficients within a prescribed neighborhood.

\subsection{Transformer layers}

A Transformer layer combines attention with a feed-forward neural network, commonly called a multilayer perceptron (MLP), together with residual connections and layer normalization. Attention combines information across token positions, while the MLP transforms each token's components independently. Residual connections add learned updates to existing representations, and normalization helps stabilize their numerical scale during training.

Stacking layers allows later queries, keys, and values to be computed from representations already transformed by earlier layers. Our probabilistic model concerns source weighting and value aggregation within attention; it does not model the complete Transformer layer or its training dynamics. 

\section{Posterior Support and Predictive Relevance}

We begin by comparing Bayesian model averaging with attention. Bayesian model averaging takes the form:

\[ P(y\mid D) = \sum_hP(h\mid D)P(y\mid h,D).  \]
Now consider attention:
\[ z_i = \sum_j\alpha_{ij}v_j.  \]

Both expressions combine candidate contributions using normalized weights. The resemblance suggests the following correspondence:

\begin{center}\begin{tabular}{p{0.43\linewidth}p{0.48\linewidth}}\hline
Bayesian inference & Transformer attention\\[3pt]
\hline
Candidate hypothesis \(h\) & Candidate contextual information source \(j\)\\[3pt]
Observed evidence \(D\) & Observed query at position \(i\)\\[3pt]
Posterior weight \(P(h\mid D)\) & Attention weight \(\alpha_{ij}\)\\[3pt]
Conditional contribution & Value vector \(v_j\)\\[3pt]
Posterior averaging & Weighted value aggregation\\[3pt]
\end{tabular}\end{center}

This structural resemblance does not automatically make attention weights Bayesian posterior probabilities. Sequential Bayesian inference accumulates evidence about hypotheses, whereas attention computes query–key scores from contextual representations. Section 6 specifies conditions under which these scores yield posterior source probabilities. Before developing those models, we distinguish support for a hypothesis from its influence on prediction.

We note that posterior support and predictive influence answer different questions. Posterior support asks how strongly the observed evidence favors a hypothesis. Predictive influence asks how changing its weight changes the predicted continuation. A highly plausible hypothesis need not strongly affect the next-token distribution if competing hypotheses make similar predictions.

To make this precise, let \(C\) denote the observed context and suppose \(H_1,H_2\) are exhaustive alternatives with \(w=P(H_1\mid C)\). Bayesian model averaging gives
\[
\begin{aligned}
P(y\mid C)
&=wP(y\mid H_1,C)+(1-w)P(y\mid H_2,C)\\
&=P(y\mid H_2,C)
+w\bigl[P(y\mid H_1,C)-P(y\mid H_2,C)\bigr].
\end{aligned}
\]
Holding the conditional predictions fixed, a change \(\Delta w\) therefore changes the predicted probability by
\[
\Delta P(y\mid C)
=\Delta w\bigl[P(y\mid H_1,C)-P(y\mid H_2,C)\bigr].
\]
Thus, sensitivity to the posterior weight depends on the difference between the hypotheses' conditional predictions. For example, if \(w=0.95\) and the two likelihoods are 0.70 and 0.69, the mixture assigns it probability \(0.95(0.70)+0.05(0.69)=0.6995\). The strong support for \(H_1\) has little effect relative to a different allocation of weights because the conditional predictions are so similar. 

When the conditional predictions differ substantially, however, a relatively unlikely hypothesis can preserve a distinctive predictive possibility. Keep \(w=0.95\), but suppose that a particular next token \(a\) has conditional probabilities
\[
P(a\mid H_1,C)=0,\qquad P(a\mid H_2,C)=0.8.
\]
The predictive mixture assigns that token probability
\[
P(a\mid C)=0.95(0)+0.05(0.8)=0.04.
\]
This shows that although \(H_2\) is relatively unlikely, it supplies the entire 4\% probability assigned to \(a\). Retaining only \(H_1\) would rule out this token. However, because a conditional probability cannot exceed one, its contribution is bounded by its posterior weight:
\[
P(H_2\mid C)P(a\mid H_2,C)
\leq P(H_2\mid C)=0.05.
\]
Thus, a small posterior weight can therefore preserve a distinctive predictive possibility without making it highly probable. This illustrates the value of averaging over explanations rather than retaining only the most plausible one, a principle also central to Solomonoff induction.

\paragraph{The corresponding distinction in attention.}
Attention separates scalar source weights from the information associated with those sources:
\[
z_i=\sum_{j\in\mathcal A_i}\alpha_{ij}v_j.
\]
A weight \(\alpha_{ij}\) determines how much source \(j\) contributes, while its value vector \(v_j\) determines what it contributes. Holding the values and other weights fixed, shifting an amount \(\delta\) of weight from source 2 to source 1 changes the output by
\[
\Delta z_i=\delta(v_1-v_2),
\]

 If \(v_1=v_2\), this redistribution changes the source weights but leaves the output unchanged. Even when the attention output changes, the final prediction need not change if the subsequent prediction function is insensitive to that change. Thus, a large attention weight alone does not establish a large influence on next-token prediction. Predictive influence depends on the values, the allocation of weights among sources, and the downstream computations. Interpreting these weights as posterior support additionally requires the probabilistic assumptions introduced in the following sections.

This motivates examining attention’s weighted value aggregation and whether the resulting representation preserves everything needed for next-token prediction, as required by predictive sufficiency, a concept discussed in Section 7

\section{Probabilistic Models of Transformer Attention}\label{sec:latent-attention}

This section develops two models of a categorical latent source \(J_i\), which identifies a candidate contextual information source. The \emph{direct conditional model} defines its probabilities using attention weights, \(P(J_i=j\mid q_i,K)=\alpha_{ij}\). Under the payload and conditioning assumptions in Section 6.3, value aggregation is a conditional expectation. No model of how the query was generated is required.

The \emph{generative model} instead specifies source priors and query likelihoods, then derives posterior source probabilities through Bayes' rule. Under the stated Gaussian and prior assumptions, these probabilities equal the attention weights, and value aggregation becomes a Bayesian posterior mean. The distinction is between specifying source probabilities directly and deriving them from a probabilistic model of the evidence.

The direct interpretation has precedents in latent-alignment and latent-connectivity work~\cite{deng2018,singh2023}; the Gaussian construction follows Gabbur et al.'s framework, including their norm-dependent prior~\cite{gabbur2021}. We develop the direct model first, followed by the generative model, then examine estimation optimality, predictive limitations, and stability. These interpretations describe what attention can compute under explicit assumptions.

\subsection{Latent variables and contextual relationships}

A latent variable is an unobserved variable introduced into a model to represent underlying states, sources, or relationships inferred from observed data. A categorical latent variable takes one of a specified set of labels. For example, consider the following sentence:
\begin{quote}The scientist placed the sample in the refrigerator because it needed cooling.\end{quote}
The pronoun ``it'' most plausibly refers to the sample, although the refrigerator is another possible interpretation. The intended referent is not explicitly annotated. We can introduce
\[
R\in\{\text{sample},\text{refrigerator}\}
\]
and ask for \(P(R=r\mid\text{sentence})\), the posterior probability of each candidate referent. This illustrates a latent contextual relationship.

\subsection{A latent source and its conditional probabilities in Transformer attention}\label{subsec:conditional-source}

For a query at position \(i\), let \(\mathcal A_i\) be the finite, nonempty set of accessible contextual source positions. Introduce a categorical latent variable
\[ J_i\in\mathcal A_i.  \]
The event \(J_i=j\) identifies position \(j\) as the information source in the latent-variable model. Each possible source therefore defines a candidate hypothesis:
\[ H_{ij}=\{J_i=j\}.  \]

\paragraph{A conditional categorical model.}
For finite query--key scores \(s_{ij}\), let
\[
P(J_i=j\mid q_i,K)
=\alpha_{ij}
=\frac{\exp(q_i^\top k_j/\sqrt{d_k})}
{\sum_{r\in\mathcal A_i}\exp(q_i^\top k_r/\sqrt{d_k})}.
\]
This is a direct probabilistic interpretation of the attention weights. These probabilities are nonnegative and sum to one over accessible sources. Here, the query and keys are conditioning inputs: we specify the categorical distribution of \(J_i\) for those inputs without modeling their probability of occurrence. 

\subsection{Value aggregation as a conditional expectation}\label{sec:value-payloads}\label{subsec:value-meaning}

At the first Transformer layer, let

\[ v_j^{(0)}=W_V^{(0)}h_j^{(0)}.  \]
This vector represents a learned projection of the information available at position \(j\). It is computed before the first layer's attention aggregation. In that limited computational sense, it is information available prior to contextual integration by the current attention head.

We view the value vector as a candidate information payload: it contains the numerical information available from source position \(j\), while the attention weight \(\alpha_{ij}\) determines how strongly that vector contributes to the attention output for query position \(i\):
\[
z_i=\sum_{j\in\mathcal A_i}\alpha_{ij}v_j.
\]
Here, ``candidate'' indicates that the source is one of the accessible positions that can contribute, and ``payload'' refers to its value vector. In deeper layers,
\[ v_j^{(\ell)} = W_V^{(\ell)}h_j^{(\ell-1)},  \]
so the value can already contain information integrated from earlier context.

\paragraph{Fixed values and a random payload.}
For a particular attention computation, the value vectors \(V=(v_1,\ldots,v_t)\) have already been computed by the Transformer, so we treat them as given, fixed information payloads in our latent-variable model. They are fixed for this computation, but can change across input contexts or during training as the learned value-projection matrix \(W_V\) and the representations it acts on change. 

Define the random payload by
\[
V_i=v_{J_i}.
\]
Although the possible value vectors are fixed, \(V_i\) is random because the source index \(J_i\) is uncertain in the model. Once the source \(J_i=j\) is known, however, the payload is fully determined: it equals \(v_j\). Thus the payload is random before the source is known, but deterministic conditional on the source and the given value vectors. The attention operation itself need not make a random selection. Transformer attention does not sample a source; it directly computes the weighted average of the value vectors

\paragraph{More general source-associated payloads.}\label{subsec:target-payload}
The payload \(V_i=v_{J_i}\), which is fully determined once the source is known, is sufficient for the direct interpretation. More generally, uncertainty about the payload may remain even after the source is known. Let \(Y_i\in\mathbb R^{d_v}\) be a random payload satisfying
\[
\mathbb E[Y_i\mid J_i=j,q_i,K,V]=v_j.
\]
The value vector represents the source-specific mean, rather than an outcome the source always supplies. Here, there are two levels of uncertainty: which source supplies the information, and what payload that source supplies. The overall mean is obtained by averaging the source-specific means using the source probabilities. This is the law of total expectation, developed explicitly below.

This generalization requires neither payloads fully determined by the source nor Gaussian payload distributions. The Gaussian payload model in Section 6.7 is one way to ensure that each source’s conditional expected payload equals its value vector \(v_j\).

\paragraph{Computing the conditional expectation.}\label{subsec:posterior-mean}

For the direct conditional model, assume that additionally conditioning on the given values does not change the specified source probabilities. The law of total expectation gives

\[ \begin{aligned} \mathbb E[Y_i\mid q_i,K,V] ={}& \sum_{j\in\mathcal A_i} P(J_i=j\mid q_i,K,V) \\ &\qquad\cdot \mathbb E[Y_i\mid J_i=j,q_i,K,V]. \end{aligned}  \]

Specifically, we assume
\[ P(J_i=j\mid q_i,K,V) = P(J_i=j\mid q_i,K). \tag{5}\label{eq:paper-31} \]

By the direct specification in Section~\ref{subsec:conditional-source},
\[ P(J_i=j\mid q_i,K)=\alpha_{ij}.  \]

By the conditional-mean assumption above,
\[ \mathbb E[Y_i\mid J_i=j,q_i,K,V] = v_j. \]

Therefore,
\[  \mathbb E[Y_i\mid q_i,K,V] = \sum_{j\in\mathcal A_i}\alpha_{ij}v_j.   \]

The right-hand side is exactly the Transformer attention output. In plain terms, we average within each source through its conditional mean and then average across sources using their probabilities. Hence,
\[  \mathbb E[Y_i\mid q_i,K,V] = z_i.   \]

\paragraph{The source-determined payload as a special case.}
For \(V_i=v_{J_i}\), which is random but fully determined once the source and value vectors are given, the same argument gives
\[
\mathbb E[V_i\mid q_i,K,V]
=\sum_{j\in\mathcal A_i}\alpha_{ij}v_j=z_i.
\]

\subsection{Source priors, query evidence, and Bayes' rule}\label{subsec:source-bayes}

We now introduce the generative interpretation. Unlike direct conditional specification, it assigns likelihoods to possible query observations under each source hypothesis and derives the source probabilities using Bayes' rule. These additional assumptions justify the weights as Bayesian posteriors. 

Before conditioning on the query, assign each source a prior probability conditional on the candidate keys \(K\):
\[ P(J_i=j\mid K)=\pi_{ij}.  \]
The prior satisfies
\[ \pi_{ij}\geq0,\qquad\sum_{j\in\mathcal A_i}\pi_{ij}=1.  \]
A simple initial choice is a uniform prior:
\[ \pi_{ij}=\frac{1}{|\mathcal A_i|}.  \]
Other choices can encode source preferences or depend on the keys. 

For the observed query \(q_i\), specify a conditional density
\[ p(q_i\mid J_i=j,K).  \]
Evaluated at that query, it gives the likelihood of each source hypothesis. We use lowercase \(p\) for densities of continuous query distributions and uppercase \(P\) for probabilities of categorical source events. Bayes' rule gives
\[ P(J_i=j\mid q_i,K)
=\frac{p(q_i\mid J_i=j,K)\pi_{ij}}
{\sum_{r\in\mathcal A_i}p(q_i\mid J_i=r,K)\pi_{ir}}. \tag{6}\label{eq:paper-39} \]

The posterior assigns probabilities to the candidate source hypotheses after accounting for the query evidence. We next establish when this posterior equals the attention weights specified in Section~\ref{subsec:conditional-source}.

\paragraph{Bayes' rule expressed through softmax.}
For positive source priors and query likelihoods, define \(\ell_{ij}=p(q_i\mid J_i=j,K)\). Rewriting Equation~\eqref{eq:paper-39} yields
\[
\begin{aligned}
P(J_i=j\mid q_i,K)
&=\frac{\pi_{ij}\ell_{ij}}{\sum_{r\in\mathcal A_i}\pi_{ir}\ell_{ir}}\\
&=\frac{\exp(\log\pi_{ij}+\log\ell_{ij})}
{\sum_{r\in\mathcal A_i}\exp(\log\pi_{ir}+\log\ell_{ir})}\\
&=\operatorname{softmax}_{j\in\mathcal A_i}
\bigl(\log\pi_{ij}+\log\ell_{ij}\bigr).
\end{aligned}
\]
Thus, softmax gives Bayesian posterior source probabilities when each score equals the log prior plus the log query likelihood, possibly
with an added constant shared by all candidate sources. This rewriting of Equation~\eqref{eq:paper-39} specifies what attention scores must represent for their softmax weights to equal the posterior probabilities under the stated model. Sections 6.5--6.6 supply a specific construction: Gaussian query likelihoods and an appropriate source prior make the log prior plus log likelihood equal the scaled query--key dot product, up to a source-independent constant.

\subsection{A Gaussian generative model and its likelihood}\label{subsec:gaussian-query}

Consider query and key vectors belonging to \(\mathbb R^{d_k}\). Conditional on the latent source variable \(J_i=j\), suppose the query is generated by
\[ q_i\mid J_i=j,K\sim\mathcal N(k_j,\sigma^2I). \tag{7}\label{eq:paper-40} \]

The key vector \(k_j\) therefore acts as the mean of the query distribution under hypothesis \(H_{ij}\). The covariance is isotropic, \(\Sigma=\sigma^2I\), with the same variance parameter \(\sigma^2\) for every candidate source. The conditional density is
\[ p(q_i\mid J_i=j,K)=\frac{1}{(2\pi\sigma^2)^{d_k/2}}\exp\left(-\frac{\|q_i-k_j\|^2}{2\sigma^2}\right). \tag{8}\label{eq:paper-41} \]
Consequently, a source whose key is closer to the observed query in Euclidean distance (a smaller \(\|q_i-k_j\|\)) assigns that query a higher likelihood. Combining this likelihood with the source’s prior probability, using Bayes' rule, gives
\[
P(J_i=j\mid q_i,K)
\propto
p(q_i\mid J_i=j,K)\,\pi_{ij}.
\]
If the priors \(\pi_{ij}\) are equal, the likelihood alone determines the ranking: the closest key receives the highest posterior probability. If the priors differ, a more distant source can receive the higher posterior probability. 

For example, suppose the query likelihoods under two sources are \(0.4\) and \(0.2\). Source 1 assigns the observed query twice the likelihood density assigned by source 2. With equal priors, their posterior probabilities are
\[
P(J_i=1\mid q_i,K)=\frac{0.4}{0.4+0.2}=\frac23,
\qquad
P(J_i=2\mid q_i,K)=\frac13.
\]
Source 1 receives greater posterior support. But if the prior probabilities are \(0.1\) and \(0.9\), the prior-weighted likelihoods become
\[
0.4(0.1)=0.04,\qquad 0.2(0.9)=0.18.
\]
After normalization, the posterior probabilities are approximately
\[
P(J_i=1\mid q_i,K)=0.182,\qquad
P(J_i=2\mid q_i,K)=0.818.
\]
Source 2 now receives greater posterior support despite its lower query likelihood.

This distinction matters for our derivation: Gaussian proximity determines likelihood, while Transformer attention corresponds to the posterior only after the source prior has also been accounted for. The following expansion makes this connection explicit.

Taking logarithms in Equation~\eqref{eq:paper-41} gives
\[ \begin{aligned}
\log p(q_i\mid J_i=j,K)
={}&-\frac{d_k}{2}\log(2\pi\sigma^2)\\
&-\frac{\|q_i-k_j\|^2}{2\sigma^2}.
\end{aligned} \tag{9}\label{eq:paper-42} \]
To make the connection to the query--key dot product explicit, expand the squared distance:
\[ \|q_i-k_j\|^2=\|q_i\|^2+\|k_j\|^2-2q_i^\top k_j.  \]
Substituting into Equation~\eqref{eq:paper-42} gives
\[ \begin{aligned}
\log p(q_i\mid J_i=j,K)
={}&-\frac{\|q_i\|^2}{2\sigma^2}
+\frac{q_i^\top k_j}{\sigma^2}\\
&-\frac{\|k_j\|^2}{2\sigma^2}+C,
\end{aligned}  \]
where \(C=-\frac{d_k}{2}\log(2\pi\sigma^2)\) is independent of the source \(j\). For a fixed query, the term involving \(\|q_i\|^2\) is also independent of \(j\). Combining these terms into \(c_i\), we obtain
\[ \log p(q_i\mid J_i=j,K)
=\frac{q_i^\top k_j}{\sigma^2}
-\frac{\|k_j\|^2}{2\sigma^2}+c_i. \tag{10}\label{eq:paper-45} \]
The term \(c_i\) depends on the query \(q_i\), but does not depend on the candidate source \(j\). It therefore contributes the same multiplicative factor \(e^{c_i}\) to every source’s likelihood. The likelihood is
\[
p(q_i\mid J_i=j,K)
=
e^{c_i}
\exp\!\left(
\frac{q_i^\top k_j}{\sigma^2}
-\frac{\|k_j\|^2}{2\sigma^2}
\right).
\]
Computing the normalized posterior, using Bayes' rule, gives
\[
P(J_i=j\mid q_i,K)
=
\frac{
\pi_{ij}e^{c_i}
\exp\!\left(
\frac{q_i^\top k_j}{\sigma^2}
-\frac{\|k_j\|^2}{2\sigma^2}
\right)
}{
e^{c_i}\displaystyle\sum_{r\in\mathcal A_i}
\pi_{ir}
\exp\!\left(
\frac{q_i^\top k_r}{\sigma^2}
-\frac{\|k_r\|^2}{2\sigma^2}
\right)
}.
\tag{11}\label{eq:normalized-gaussian-posterior}
\]
The common factor \(e^{c_i}\) cancels out, and since the denominator is independent of \(j\), we can write:
\[ P(J_i=j\mid q_i,K) = c\times\pi_{ij} \exp \left( \frac{q_i^\top k_j}{\sigma^2} - \frac{\|k_j\|^2}{2\sigma^2} \right), \]
where \(c\) is the reciprocal of the remaining normalization denominator. It is constant across candidate sources \(j\), though it depends on the given query. We can now state 
\[ P(J_i=j\mid q_i,K) \propto \pi_{ij} \exp \left( \frac{q_i^\top k_j}{\sigma^2} - \frac{\|k_j\|^2}{2\sigma^2} \right). \tag{12}\label{eq:paper-46} \]
That is, only source-dependent terms influence their relative posterior probabilities.

Equation~\eqref{eq:paper-45} shows that the Gaussian log likelihood depends on both the query--key dot product and a key-norm penalty. If all candidate keys have equal norms, the penalty is constant across sources, so minimizing Euclidean distance is equivalent to maximizing the dot product. With unequal key norms, proximity and dot-product rankings need not agree.

Write each prior as
\(\pi_{ir}=\exp(\log\pi_{ir})\).
Applying this identity in both the numerator and denominator of
Equation~\eqref{eq:normalized-gaussian-posterior}, and cancelling
the common factor \(e^{c_i}\), expresses the posterior as a softmax
with scores
\[
b_{ij}
=\frac{q_i^\top k_j}{\sigma^2}
-\frac{\|k_j\|^2}{2\sigma^2}
+\log\pi_{ij}.
\tag{13}\label{eq:paper-47}
\]
The posterior therefore becomes
\[
P(J_i=j\mid q_i,K)
=\operatorname{softmax}_{j\in\mathcal A_i}(b_{ij}).
\tag{14}\label{eq:paper-48}
\]

This is a general Bayesian form of attention under the Gaussian model. Equation~\eqref{eq:paper-47} contains three source-dependent contributions: query--key compatibility, a key-norm penalty from the Gaussian likelihood, and the log prior. For uniform priors and unequal key norms, the penalty remains, so the posterior generally differs from ordinary dot-product attention. The following subsection shows how equal key norms with uniform priors, or a norm-dependent prior for unequal key norms, recover standard scaled dot-product attention.

\subsection{Recovering standard scaled dot-product attention}\label{subsec:recover-attention}

We first consider a simple special case. Assume that the prior probabilities \(\pi_{ij}\) are uniform, all candidate keys have equal Euclidean norm, and the variance parameter satisfies 
\[ \sigma^2=\sqrt{d_k}. \tag{15}\label{eq:paper-49} \]

Because the priors are uniform, \(\log\pi_{ij}\) is constant across \(j\). Because the key norms are equal, the term \( -\frac{\|k_j\|^2}{2\sigma^2} \) is also constant across \(j\). And since both terms disappear under softmax normalization, Equation~\eqref{eq:paper-48} therefore reduces to
\[  P(J_i=j\mid q_i,K) = \frac{ \exp(q_i^\top k_j/\sqrt{d_k}) }{ \sum_{r\in\mathcal A_i} \exp(q_i^\top k_r/\sqrt{d_k}) }.  \tag{16}\label{eq:paper-50} \]
But the right-hand side is precisely the standard attention weight \( \alpha_{ij} \).

An exact correspondence can also be constructed without equal key norms. For any finite, nonempty accessible source set and finite key vectors, choose the conditional source prior
\[
\pi_{ij}=P(J_i=j\mid K)
=\frac{\exp(\|k_j\|^2/(2\sigma^2))}
{\sum_{r\in\mathcal A_i}\exp(\|k_r\|^2/(2\sigma^2))}.
\]
This is a valid prior over accessible sources conditional on the keys: it is positive, sums to one, and does not depend on the observed query. Let
\[
C_i(K)=\sum_{r\in\mathcal A_i}\exp(\|k_r\|^2/(2\sigma^2)).
\]
Substituting the logarithm of \(\pi_{ij}\) into Equation~\eqref{eq:paper-47} gives
\[
\begin{aligned}
b_{ij}
&=\frac{q_i^\top k_j}{\sigma^2}
-\frac{\|k_j\|^2}{2\sigma^2}
+\frac{\|k_j\|^2}{2\sigma^2}-\log C_i(K)\\
&=\frac{q_i^\top k_j}{\sigma^2}-\log C_i(K).
\end{aligned}
\]
The key-norm terms cancel, and the remaining term \(-\log C_i(K)\) is constant across candidate sources. Therefore,
\[
P(J_i=j\mid q_i,K)
=\operatorname{softmax}_{j\in\mathcal A_i}
\left(\frac{q_i^\top k_j}{\sigma^2}\right).
\]
With Equation~\eqref{eq:paper-49}, this is again exactly Equation~\eqref{eq:paper-50}, now for arbitrary key norms. When the keys have equal norms, the constructed prior becomes uniform, \(\pi_{ij}=1/|\mathcal A_i|\), recovering the earlier special case.

We can now state the correspondence for arbitrary key norms, with the uniform-prior, equal-norm construction included as a special case.

\subsection*{Proposition 1. Bayesian interpretation of scaled dot-product attention}

Consider the Gaussian latent-source model of Equation~\eqref{eq:paper-40} on a finite, nonempty accessible source set \(\mathcal A_i\), with finite key vectors and \(\sigma^2=\sqrt{d_k}\). Suppose the source prior satisfies
\[
\log\pi_{ij}-\frac{\|k_j\|^2}{2\sigma^2}=c_i(K)
\qquad\text{for all }j\in\mathcal A_i,
\]
where \(c_i(K)\) is independent of \(j\). Equivalently, choose
\[
\pi_{ij}
=\frac{\exp(\|k_j\|^2/(2\sigma^2))}
{\sum_{r\in\mathcal A_i}\exp(\|k_r\|^2/(2\sigma^2))}.
\]
Then the standard scaled dot-product attention weights are exactly the posterior probabilities of the latent source variable:
\[  P(J_i=j\mid q_i,K) = \alpha_{ij}.  \]

\subsection{Value aggregation as a Bayesian posterior mean}
We now combine Proposition 1 with the conditional-mean assumptions of Section 6.3 to interpret value aggregation as a Bayesian posterior mean.

\paragraph{What is being averaged.}
Each value vector represents a conditional payload mean:
\[
v_j=\mathbb E[Y_i\mid J_i=j,q_i,K,V].
\]
The information supplied by source \(j\) need not always equal \(v_j\); the payload may vary while retaining that average. Under Proposition 1 and the conditioning assumption in Equation~\eqref{eq:paper-31}, \(P(J_i=j\mid q_i,K,V)=\alpha_{ij}\). The law of total expectation first averages conditional payload means over the possible sources:
\[
\begin{aligned}
\mathbb E[Y_i\mid q_i,K,V]
&=\sum_{j\in\mathcal A_i}P(J_i=j\mid q_i,K,V)\,
\mathbb E[Y_i\mid J_i=j,q_i,K,V]\\
&=\sum_{j\in\mathcal A_i}\alpha_{ij}v_j\\
&=z_i.
\end{aligned}
\]
Hence
\[
\underbrace{\sum_{j\in\mathcal A_i}\alpha_{ij}v_j}_{\text{attention output}}
=\underbrace{\mathbb E[Y_i\mid q_i,K,V]}_{\text{posterior mean}}.
\]
Attention therefore combines the source-specific means according to how probable each source is after observing the query.

\paragraph{Posterior mean, MAP estimation, and EM.}
Our analysis emphasizes posterior-mean estimation, complementing Gabbur et al.'s treament of Maximum A Posteriori (MAP) inference and Expectation Maximization (EM) based adaptation~\cite{gabbur2021}. A posterior mean averages possible payloads, whereas MAP identifies a most probable value; EM is an estimation algorithm rather than an alternative posterior summary. Our focus is on when the resulting mean preserves information needed for prediction. The conditional-mean assumption is sufficient for this interpretation; the Gaussian payload model below provides one concrete realization.

\paragraph{A Gaussian realization.}
Here we specify a concrete Gaussian payload model for our latent-source analysis, drawing on the probabilistic-attention perspective of Gabbur et al.~\cite{gabbur2021}. This serves two purposes. First, it provides a model with the conditional payload means required for posterior-mean aggregation. Second, it ensures the finite conditional second moment needed for the squared-error optimality argument in the subsequent section.

The Gaussian query distribution in Equation~\eqref{eq:paper-40}, centered at \(k_j\), supplies the likelihood used to infer the source. A second Gaussian distribution, centered at \(v_j\), models the payload associated with that source:
\[ Y_i\mid J_i=j,K,V\sim\mathcal N(v_j,\tau^2I). \tag{17}\label{eq:paper-52} \]
Here \(V=(v_1,\ldots,v_t)\). This distribution gives \(\mathbb E[Y_i\mid J_i=j,K,V]=v_j\), but observing the query could change that mean. For example, a source may have mean payload \(5\), while the query reveals whether the current payload is above or below \(5\). To rule out such additional information in this realization, assume conditional independence:
\[ Y_i\perp q_i\mid J_i,K,V. \tag{18}\label{eq:paper-53} \]
Together, these assumptions give
\[ \mathbb E[Y_i\mid J_i=j,q_i,K,V]=v_j. \tag{19}\label{eq:paper-54} \]
The query can therefore change our probabilities over sources without changing the payload distribution once the source, keys, and values are known. This conditional independence is sufficient but stronger than necessary: the posterior-mean identity requires only that the conditional mean remain \(v_j\), as in Equation~\eqref{eq:paper-54}. Other aspects of the payload distribution could depend on the query without invalidating that identity. Thus, the query likelihood supports source inference, while the payload distribution specifies the information to be estimated.

This Gaussian realization also ensures the finite conditional second moment needed for the squared-error optimality argument in Section 6.8. Here, \(\mathbb E[|Y_i|^2\mid q_i,K,V]\) is the expected squared magnitude of the payload given the query, keys, and values.

More generally, a Gaussian distribution for the payload is sufficient but not necessary; the posterior-mean identity requires only the conditional-mean and conditioning assumptions established in Section 6.3.

\subsection*{Corollary 1. Attention as a Bayesian posterior mean}

Consider the Gaussian latent-source model defined by Equations~\eqref{eq:paper-40}, \eqref{eq:paper-52}, and \eqref{eq:paper-53}, on a finite, nonempty accessible source set, with finite keys and values. Suppose \(\sigma^2=\sqrt{d_k}\) and the source prior is the norm-dependent prior specified in Proposition 1. Treat the values as fixed conditional payload parameters, so that the source posterior is unchanged by additionally conditioning on \(V\), as in Equation~\eqref{eq:paper-31}. No equal-key-norm assumption is required; uniform priors with equal key norms are a special case.

Then scaled dot-product attention computes the posterior conditional expectation of the target contextual information vector:

\[ \operatorname{Attention}(q_i,K,V) = \mathbb E[Y_i\mid q_i,K,V].   \]

This follows directly from Proposition 1 and the conditional-expectation calculation in Section~\ref{sec:value-payloads}, under the payload and conditioning assumptions stated above.

\subsection{Mean-estimation optimality and limits for prediction}\label{subsec:mean-prediction}

Let \(a\in\mathbb R^{d_v}\) be any estimator of \(Y_i\) based on the available evidence. Define the conditional risk; that is, the expected squared estimation error given the query, keys, and values, by
\[ \mathcal R(a) = \mathbb E[ \|Y_i-a\|^2 \mid q_i,K,V ].  \]
Given this query, these keys, and these values, how much squared error should we expect if we estimate the payload by \(a\)?

Let
\[ m_i=\mathbb E[Y_i\mid q_i,K,V]. \]

Using
\[ Y_i-a=(Y_i-m_i)+(m_i-a) \]

and expanding, we obtain
\[ \begin{aligned} \mathcal R(a) ={}& \mathbb E[\|Y_i-m_i\|^2\mid q_i,K,V] \\ &+\|m_i-a\|^2 \\ &+2\mathbb E[(Y_i-m_i)^\top(m_i-a)\mid q_i,K,V]. \end{aligned}  \]

Because
\[ \mathbb E[Y_i-m_i\mid q_i,K,V]=0, \]

We have
\[ \mathcal R(a) = \mathcal R(m_i) + \|m_i-a\|^2.   \]

The risk is minimized when
\[ a=m_i. \]

Section~\ref{sec:value-payloads} establishes that the attention output equals the conditional mean; Corollary 1 gives its posterior-mean interpretation in the generative case. Therefore, the optimal squared-error estimate is
\[  a^*=z_i.   \]

Thus, under our probabilistic model, the attention output is not merely a weighted vector average. It is the optimal conditional-mean estimator of the modeled payload under squared-error loss; under the generative assumptions, it is the corresponding Bayes-optimal estimator.

\paragraph{From payload estimation to next-token prediction.}
This optimality concerns estimation of the modeled payload, not automatically prediction of the next token. If \(Y_i\) is a contextual information vector, the result establishes the best squared-error estimate of that vector given the query, keys, and values. It does not establish that its mean preserves all information needed for next-token prediction. Its predictive relevance therefore depends on what the payload represents, what predictive distinctions its mean preserves, and how subsequent computations use that mean.

A posterior mean can nevertheless preserve a complete predictive distribution when the payload and source probabilities have the appropriate interpretation. Section 7.3 gives an illustrative construction using source-conditional next-token distributions.

\paragraph{Predicting from an average is not the same as averaging predictions.}
A further distinction, emphasized by Deng et al.~\cite{deng2018}, concerns the computation after attention. Let \(G\) be a downstream prediction function, let \(C\) denote the conditioning context, and let \(J\) follow the conditional source probabilities. In general,
\[
G\!\left(\mathbb E[v_J\mid C]\right)
\ne \mathbb E[G(v_J)\mid C].
\]
The left-hand side applies the prediction function to the mean representation. This is the order used by ordinary Transformer attention: first average value vectors, then process the aggregated representations through subsequent computations to obtain outcome probabilities. Thus \(G(z_i)\) denotes a prediction from an average representation, not an average of source-specific outcome probabilities. 

Bayesian predictive averaging instead combines conditional outcome distributions using posterior hypothesis probabilities. For a categorical source \(J\),
\[
P(X_{t+1}=a\mid C)
=\sum_jP(J=j\mid C)P(X_{t+1}=a\mid J=j,C).
\]
This can be described as ``apply \(G\) first, then average,'' provided \(G(v_j)\) gives the source-conditional outcome distribution:
\[
[G(v_j)]_a=P(X_{t+1}=a\mid J=j,C)
\quad\text{for every token }a.
\]
Under this assumption, and with the corresponding posterior source weights from the same model, the predictive distribution is
\[
\sum_jP(J=j\mid C)G(v_j)
=\mathbb E[G(v_J)\mid C].
\]
Each source is first assigned a predictive distribution, and these distributions are then averaged using posterior weights. The quantities being averaged are probabilities of the same outcomes under different source hypotheses. This distinguishes the operation from the Transformer order of averaging value vectors before applying the downstream prediction function.

The computations coincide in the special construction where each value vector already encodes the conditional next-token distribution and the attention weights are its source posterior probabilities. Then \(z_i\) itself is the full predictive distribution and \(G\) can be the identity.

For a simple numerical example, take two equally probable scalar values,
\[
v_1=-1,\qquad v_2=1,
\]
and the nonlinear transformation \(G(v)=v^2\). Averaging first gives
\[
\mathbb E[v_J\mid C]=\tfrac12(-1)+\tfrac12(1)=0,
\qquad
G\!\left(\mathbb E[v_J\mid C]\right)=0.
\]
Applying the transformation first instead gives
\[
\mathbb E[G(v_J)\mid C]
=\tfrac12(-1)^2+\tfrac12(1)^2=1.
\]
Second-moment information matters here specifically because \(G(v)=v^2\): averaging the transformed values gives \(\mathbb E[v_J^2\mid C]\), the conditional second raw moment, whereas transforming the mean gives \((\mathbb E[v_J\mid C])^2\). Their difference is the conditional variance:
\[
\mathbb E[v_J^2\mid C]-(\mathbb E[v_J\mid C])^2
=\operatorname{Var}(v_J\mid C).
\]
In this example the mean is \(0\), the second moment is \(1\), and the variance is \(1\). The mean alone does not determine the second moment needed to reproduce the average of squared values. For other nonlinear functions, the average prediction may depend on other features of the distribution beyond its mean.

\paragraph{Why concatenating heads does not automatically resolve the distinction.}
Each attention head computes an output for every query position. At a fixed position \(i\), the outputs of the different heads are concatenated. Writing each head output as a column vector, the combined representation before output projection is
\[
s_i=\begin{bmatrix}z_i^{(1)}\\ \vdots\\ z_i^{(m)}\end{bmatrix},
\]
with projected output \(W_O^\top s_i\) under the matrix convention of Section 4.5. The full Transformer subsequently applies further computations, including nonlinear transformations.

The probabilistic result for each head is conditional: \(z_i^{(r)}=\mathbb E[Y_i^{(r)}\mid q_i^{(r)},K^{(r)},V^{(r)}]\). To interpret all concatenated outputs as one joint posterior mean, a compatible joint model and common conditioning information \(C\) are needed. If such a model satisfies \(z_i^{(r)}=\mathbb E[Y_i^{(r)}\mid C]\) for every head, then stacking the payloads into \(\mathbf Y_i\) gives \(s_i=\mathbb E[\mathbf Y_i\mid C]\) by linearity. 

Even under this common-conditioning assumption, a nonlinear downstream prediction function \(G\), including the output projection and subsequent computations as appropriate, need not satisfy
\[
G(s_i)=G\!\left(\mathbb E[\mathbf Y_i\mid C]\right)
=\mathbb E[G(\mathbf Y_i)\mid C].
\]
The first equality follows from the mean representation; the second is not guaranteed. Concatenation preserves the individual head outputs, and the output projection is linear, but neither operation makes later nonlinear prediction make predicting from the averaged representations equivalent to averaging predictions from the individual representations. Additional heads can retain complementary information, yet no architectural guarantee ensures that their means preserve every distinction needed to reproduce a predictive mixture.

This is not to say that a Transformer should perform source-specific predictive averaging, or that its actual predictions are incorrect. This observation establishes a limit on the conclusion drawn from the posterior-mean identity: correct mean estimation alone does not prove Bayesian predictive marginalization or next-token optimality. Whether the combined representation contains the information needed for the prediction task is the predictive-sufficiency question developed in Section 7.

\subsection{Stability under approximate source weighting}\label{subsec:source-stability}

We now quantify how errors in source probabilities affect the conditional or posterior mean, with the value vectors held fixed. Fix the query, keys, and value vectors in the generative latent-source model of Sections 6.4–6.7. Let \(p\) be the true posterior distribution over accessible sources and \(\widehat p\) an approximation to it. We examine how replacing \(p\) by \(\widehat p\), while keeping the value vectors unchanged, affects the aggregated output.

Let the true posterior source probabilities be
\[ p_j=P(J_i=j\mid q_i,K), \]

and let their attention-based approximations be
\[ \widehat p_j. \]

Suppose the value vectors satisfy
\[ \|v_j\|\leq B \qquad\text{for every }j. \tag{20}\label{eq:paper-60} \]

The true posterior mean is
\[ m=\sum_jp_jv_j.  \]

The approximate attention output is
\[ \widehat m=\sum_j\widehat p_jv_j.  \]

We can bound the resulting approximation error as follows.

\subsection*{Proposition 2. Stability of attention-based posterior expectations}

Under Equation~\eqref{eq:paper-60},
\[  \|\widehat m-m\| \leq 2B\,D_{\mathrm{TV}}(p,\widehat p),   \]

Here, total variation distance \(D_{\mathrm{TV}}(p,\widehat p)\) measures the largest absolute difference between the probabilities that the two distributions assign to the same event. The proof first bounds the representation error using total variation; Pinsker's inequality then converts this into a bound using KL divergence.

\subsection*{Proof}

Subtract the two representations:
\[ \widehat m-m = \sum_j(\widehat p_j-p_j)v_j.  \]

Apply the triangle inequality:
\[ \|\widehat m-m\| \leq \sum_j|\widehat p_j-p_j|\|v_j\|.  \]

Using \(\|v_j\|\leq B\),
\[ \|\widehat m-m\| \leq B\sum_j|\widehat p_j-p_j|.  \]

Using the standard identity for total variation distance~\cite{levinPeres2017},
\[ D_{\mathrm{TV}}(p,\widehat p) = \frac12\sum_j|p_j-\widehat p_j|,  \]

we obtain
\[ \|\widehat m-m\| \leq 2B\ D_{\mathrm{TV}}(p,\widehat p). \]

This proves the proposition. \(\square\)

Using natural logarithms, Kullback–Leibler divergence is defined by
\[
D_{\mathrm{KL}}(p\|\widehat p)
=\sum_{j:p_j>0}p_j\ln\frac{p_j}{\widehat p_j}.
\]
It computes the expected additional log loss from using the approximate source probabilities \(\widehat p\) when the true source probabilities are \(p\). The divergence is zero when the distributions agree and positive otherwise. If \(\widehat p_j=0\) for a source with \(p_j>0\), the divergence is infinite.

Pinsker's inequality~\cite{coverThomas2006}, expressed using natural logarithms, gives
\[
D_{\mathrm{TV}}(p,\widehat p)
\leq\sqrt{\frac12D_{\mathrm{KL}}(p\|\widehat p)}.
\]
Combining this with Proposition 2 yields
\[
\|\widehat m-m\|\leq B\sqrt{2D_{\mathrm{KL}}(p\|\widehat p)}.
\]

This establishes a direct connection between the quality of posterior source approximation and the accuracy of the resulting attention representation. It connects errors in source inference to errors in the attention output. In plain terms, a small KL divergence between the true and approximate source posteriors guarantees a small error in their weighted value means, provided the same value vectors are used and their norms are bounded by \(B\). This is an upper bound, not an exact error estimate.

During pretraining, gradient-based optimization changes the projection matrices and other parameters to reduce next-token loss. During ordinary inference, these parameters remain fixed, while contextual representations and attention weights vary with the input. This input dependence permits different source contributions across contexts, but does not by itself establish that the weights accurately represent posterior probabilities over contextual sources or that the aggregated values preserve all predictive information.

\section{Predictive Sufficiency and Information Compression}

Section 6 establishes conditions under which attention computes a posterior mean of a contextual payload. This answers an estimation question: what vector should represent the uncertain information supplied by the candidate sources? It does not yet answer the prediction question: does that vector contain everything needed to predict the next token?

The distinction matters because averaging combines several possibilities into one vector. That vector can be exactly the right estimate of the modeled payload while omitting distinctions relevant to another task. Conversely, an average can retain an entire predictive distribution when its components encode the appropriate probabilities. The issue is therefore not whether averaging is inherently adequate or inadequate, but what is averaged and what must subsequently be predicted.

This section develops that distinction in three steps. First, we introduce predictive sufficiency and illustrate how a short summary can preserve all information needed for prediction. Second, we examine when attention produces such a summary and separate representation information loss from prediction-head error. Third, we connect this analysis to information compression and algorithmic prediction bounds, identifying the additional approximation assumption needed for the latter to bound representation loss.

Preserving information needed for prediction is an established concern in the information bottleneck literature~\cite{tishby1999} and the information-sufficiency framework of Silva et al.~\cite{silva2024}. Related work examines attention identifiability~\cite{brunner2020} and the distinction between representation averaging and predictive marginalization~\cite{deng2018}. Our contribution is to apply these perspectives to the posterior-mean interpretation of attention: when does an attention-weighted mean preserve the information needed for next-token prediction, and when can averaging discard relevant distinctions?

\subsection{Sufficient statistics}

In statistics, a statistic is a function of the observed data:
\[ S=T(X). \]
The classical concept of statistical sufficiency, associated with R. A. Fisher~\cite{fisher1922}, concerns retaining the information in observations relevant to an unknown parameter. Under a specified statistical model, \(S\) is sufficient for a parameter \(\theta\) if the conditional distribution of \(X\) given \(S\) does not depend on \(\theta\). Once the statistic is known, the remaining details of the observations provide no additional information about that parameter. 

A statistic \(S=T(X)\) can therefore be viewed as a summary of the observations. Classical sufficiency asks whether this summary preserves everything relevant to inference about a parameter. Predictive sufficiency asks a related but different question: does the summary retain everything needed to predict a specified target? Here, we apply it to attention-based representations for next-token prediction. 

A representation
\[ S_t=\Phi(X_{1:t}) \]
is predictively sufficient when
\[ P(X_{t+1}\mid X_{1:t})=P(X_{t+1}\mid S_t). \]
The equality holds for every vocabulary token in every context that has positive probability under the data-generating distribution. Once the representation is known, access to the full context provides no further predictive information. The summary need not reconstruct the original input sequence or retain every fact in it; it must retain the distinctions that affect the next-token distribution. Thus, the distinction is the purpose of the summary. Our contribution is to apply predictive sufficiency to attention-based representations and ask whether posterior-mean aggregation preserves the distinctions needed for next-token prediction.

A useful way to recognize failure is to consider two different input sequences mapped to the same representation. Any prediction head using only that representation must give them the same prediction. If their true next-token distributions differ, the representation cannot be sufficient for both. In probabilistic terms, predictive sufficiency says that the next token and full context are conditionally independent given the summary. 

\subsection{An example: Bayesian inference without storing the entire sequence}

Consider a sequence over the alphabet \(\{A,B\}\). One of two hypotheses generates the entire sequence, but we do not know which one was chosen. Before observing any symbols, the hypotheses have equal prior probabilities:
\[
P(H_1)=P(H_2)=\tfrac12.
\]

Suppose that their conditional symbol probabilities are
\[
H_1:\quad P(A\mid H_1)=0.8,\qquad P(B\mid H_1)=0.2,
\]
\[
H_2:\quad P(A\mid H_2)=0.2,\qquad P(B\mid H_2)=0.8.
\]

An intuitive analogy is to choose one of two biased coins at random and repeatedly toss that same coin. The first favors \(A\), while the second favors \(B\).

The observations are independent conditional on the chosen hypothesis: if we knew the generating rule, earlier outcomes would not change the probability of the next symbol. Thus,
\[
P(AAB\mid H_1)=0.8\times0.8\times0.2=0.128,
\qquad
P(AAB\mid H_2)=0.2\times0.2\times0.8=0.032.
\]
The observed sequence \(AAB\) is four times as likely under \(H_1\). Bayes' rule, with equal priors, gives
\[
P(H_1\mid AAB)
=\frac{(1/2)(0.128)}{(1/2)(0.128)+(1/2)(0.032)}=0.8,
\qquad P(H_2\mid AAB)=0.2.
\]
The next-symbol prediction averages the two rules using these posterior probabilities:
\[
P(X_4=A\mid AAB)=0.8(0.8)+0.2(0.2)=0.68.
\]
Although observations are independent when the hypothesis is known, they are not independent when it is unknown. For example, before observing any symbols, our predictive probability of \(A\), averaged over the two equally likely hypotheses, is \(0.5\), but observing one \(A\) raises the posterior probability of \(H_1\) to \(0.8\) and the probability of another \(A\) to \(0.68\).  Neither generating rule assigns \(A\) probability \(0.5\): one assigns \(0.8\), the other \(0.2\). The value \(0.5\) expresses our uncertainty about which rule was selected. The observation changes our knowledge of the fixed generating rule, rather than changing that rule.

We now show why the complete history is unnecessary for these calculations. Let
\[ N_A(t)=\sum_{i=1}^{t}\mathbf1\{X_i=A\}, \]
\[ N_B(t)=t-N_A(t). \]

With equal hypothesis priors, the posterior odds are

\[ \frac{P(H_1\mid X_{1:t})} {P(H_2\mid X_{1:t})} = \frac{ 0.8^{N_A(t)}0.2^{N_B(t)} }{ 0.2^{N_A(t)}0.8^{N_B(t)} }.  \]

Taking logarithms gives

\[  \log \frac{P(H_1\mid X_{1:t})} {P(H_2\mid X_{1:t})} = (N_A(t)-N_B(t))\log4.   \]

Thus, the complete posterior can be recovered from the single number

\[ S_t=N_A(t)-N_B(t).  \]

Writing \(s=N_A(t)-N_B(t)\), the posterior probability of \(H_1\) is \(4^s/(1+4^s)\). Consequently,
\[
P(X_{t+1}=A\mid X_{1:t})
=0.8\frac{4^s}{1+4^s}+0.2\frac{1}{1+4^s}.
\]
For the sequence \(AAB\), \(s=2-1=1\), recovering the next-symbol probability \(0.68\) calculated above. Every history with that same count difference gives the same prediction in this model, regardless of its order. The complete sequence and the series of earlier posteriors are unnecessary for this prediction. Thus, \(S_t=N_A(t)-N_B(t)\) is a predictively sufficient summary for the next symbol under this model: it needs not preserve the whole context, only what affects the predictive distribution. 

\subsection{From attention to predictive summaries}

With learned parameters \(\theta\) fixed, write the Transformer's representation and predictor as
\[
S_t^\theta=\Phi_\theta(X_{1:t}),\qquad
\widehat P_\theta(\cdot\mid X_{1:t})=G_\theta(S_t^\theta),
\]
where \(\Phi_\theta\) includes attention layers and \(G_\theta\) outputs a vocabulary distribution. Predictive sufficiency concerns what the representation preserves; prediction accuracy also depends on how the trained head uses it. For the combined representation and head, an accuracy requirement is
\[ D_{\mathrm{KL}} \left( P^*(\cdot\mid X_{1:t}) \middle\| G_\theta(S_t^\theta) \right) \leq\varepsilon, \tag{21}\label{eq:paper-78} \]
where \(P^*\) is the true next-token distribution and \(\varepsilon\) a specified tolerance. Equation~\eqref{eq:paper-78} controls both components; the next subsection separates excess log loss into contributions from information lost by the representation and error introduced by the prediction head.

\paragraph{Example: When the posterior mean gives the full next-token distribution.}
Corollary 1 identifies attention as the posterior mean of a general contextual payload, without establishing predictive sufficiency. Here we add a specific assumption: values encode source-conditional next-token distributions in a correctly specified model.

Let \(C=X_{1:t}\) be the context, \(J\) a categorical latent source with finite source set \(\mathcal A\), and \(\mathcal X\) the finite vocabulary. Suppose
\[
v_j(a;C)=P(X_{t+1}=a\mid J=j,C),\qquad
\alpha_j(C)=P(J=j\mid C).
\]
The component \(v_j(a;C)\) corresponds to token \(a\); the full vector \(v_j(\cdot;C)\) has nonnegative components summing to one. The law of total probability gives
\[
z(a;C)=\sum_{j\in\mathcal A}\alpha_j(C)v_j(a;C)
=P(X_{t+1}=a\mid C),\qquad a\in\mathcal X.
\]
Under these additional assumptions, the posterior-weighted mean \(z(\cdot;C)\) contains the full next-token distribution and is predictively sufficient; an identity prediction head suffices.

For example, two sources with values
\[
v_1=(0.8,0.1,0.1)^\top,\qquad v_2=(0.2,0.3,0.5)^\top
\]
and posterior weights \(0.75\) and \(0.25\) produce
\[
z=0.75v_1+0.25v_2=(0.65,0.15,0.20)^\top.
\]
All three token probabilities are retained.

To connect this construction to conditional-mean estimation, let \(T\) be the random one-hot indicator of the next token: \(T_a=1\) if token \(a\) occurs and \(0\) otherwise. Unlike \(T\), \(V=(v_j(\cdot;C))_{j\in\mathcal A}\) is the collection of value vectors, given for a fixed context and fixed parameters. A realization of \(T\) identifies one outcome; each value vector describes probabilities over outcomes under a source hypothesis. Consequently,
\[
v_j(\cdot;C)=\mathbb E[T\mid J=j,C],\qquad
z(\cdot;C)=\mathbb E[T\mid C].
\]
Each component's mean equals its token's probability because its indicator is \(1\) exactly when that token occurs. Thus the example specializes the earlier general payload \(Y_i\) to a prediction-specific payload \(T\).

The example also replaces conditioning on \(q_i,K,V\) with conditioning on the full context \(C\). Applying it to ordinary attention would require the contextual token representations \(h_j\) to retain enough information, and the learned projections producing \(q_i\), \(k_j\), and \(v_j\) to use that information correctly, so that the attention weights and value vectors encode the source probabilities and conditional next-token distributions assumed in the example.

\subsection{A decomposition of excess predictive log loss}

Fix a data-generating distribution and a deterministic representation \(S_t=\Phi(X_{1:t})\), with learned parameters held fixed. Define
\[
\begin{aligned}
p_t^*(\cdot)&=P(X_{t+1}=\cdot\mid X_{1:t}),\\
p_t^S(\cdot)&=P(X_{t+1}=\cdot\mid S_t),\\
\widehat p_t(\cdot)&=G_\theta(S_t).
\end{aligned}
\]
These are, respectively, the ideal full-context predictor, the ideal representation-based predictor, and the trained head's prediction. The first two minimize expected log loss given their available information. The head receives only \(S_t\).

Using a predictive distribution \(q\) incurs log loss \(-\ln q(a)\) when token \(a\) occurs. If the true conditional distribution is \(p\), the expected additional log loss from using \(q\) rather than \(p\) is \(D_{\mathrm{KL}}(p\|q)\). Throughout this subsection, we use natural logarithms.

\paragraph{Two levels of averaging.}
For a fixed context \(c\), let \(p_c\) and \(\widehat p_c\) be its true and predicted next-token distributions. KL divergence averages over next tokens; the outer expectation then averages over contexts:
\[
\mathbb E[D_{\mathrm{KL}}(p_t^*\|\widehat p_t)]
=\sum_c P(X_{1:t}=c)\sum_{a:p_c(a)>0}p_c(a)
\ln\frac{p_c(a)}{\widehat p_c(a)}.
\]
The following decomposition, presented by  Silva et al.~\cite{silva2024}, separates the two sources of excess loss:
\[  \begin{aligned} &\mathbb E[ D_{\mathrm{KL}}(p_t^*\|\widehat p_t) ] \\ &\quad= I(X_{t+1};X_{1:t}\mid S_t) + \mathbb E[ D_{\mathrm{KL}}(p_t^S\|\widehat p_t) ]. \end{aligned}  \tag{22}\label{eq:paper-79} \]

Here, we apply it with the full context \(X_{1:t}\) as input, the next token \(X_{t+1}\) as target, and the attention-based representation \(S_t\) as summary.

Conditional mutual information measures how much the full context still helps predict the next token once the summary is known. In this setting,
\[
I(X_{t+1};X_{1:t}\mid S_t)
=\mathbb E[D_{\mathrm{KL}}(p_t^*\|p_t^S)].
\]
It is the expected additional log loss from using the ideal summary-based predictor instead of the ideal full-context predictor, averaged over contexts.

The first term in Equation~\eqref{eq:paper-79} is \emph{representation information loss}: the uncertainty that access to the full context would remove after the summary is known. It vanishes exactly when the summary is predictively sufficient. The second is \emph{prediction-head error}: the head's discrepancy from the best predictor available from that summary. A sufficient representation can still have a poor head, while a perfect head cannot recover information absent from its input.

\paragraph{Example: separating the two costs.}
Suppose two equally likely contexts have next-token distributions \(p_1=(0.9,0.1)\) and \(p_2=(0.1,0.9)\), but map to the same summary. Its ideal predictor must average the two distributions:
\[
p^S=\tfrac12p_1+\tfrac12p_2=(0.5,0.5).
\]
Representation loss averages the divergence from each full-context distribution to this summary-based predictor:
\[
I(X_{t+1};X_{1:t}\mid S_t)
=\tfrac12D_{\mathrm{KL}}(p_1\|p^S)
+\tfrac12D_{\mathrm{KL}}(p_2\|p^S).
\]
For the first context,
\[
\begin{aligned}
D_{\mathrm{KL}}(p_1\|p^S)
&=0.9\ln\frac{0.9}{0.5}+0.1\ln\frac{0.1}{0.5}\\
&=0.9\ln(1.8)+0.1\ln(0.2)\\
&\approx0.368\text{ nats}.
\end{aligned}
\]
For the second context, the two contributions appear in the opposite order:
\[
D_{\mathrm{KL}}(p_2\|p^S)
=0.1\ln(0.2)+0.9\ln(1.8)
=D_{\mathrm{KL}}(p_1\|p^S).
\]
The divergences are equal because the contexts exchange the token probabilities while the summary-based predictor assigns both tokens the same probability. Their equally weighted average is therefore also approximately \(0.368\) nats. This is the additional expected log loss caused by losing the contextual distinction, even with an ideal prediction head.

If the trained head instead outputs \((0.6,0.4)\), it adds
\[
D_{\mathrm{KL}}((0.5,0.5)\|(0.6,0.4))\approx0.0204\text{ nats},
\]
for a total excess loss of approximately \(0.388\) nats. Improving the head removes the second cost; preserving the distinction between contexts is needed to remove the first.

\paragraph{Sensitivity of predictions to source-weight errors.}\label{par:prediction-stability}
The loss decomposition concerns predictive information loss and head error. The following consequence of Proposition 2 instead bounds how a fixed head's output changes when its input is formed with approximate source weights.

\paragraph{Corollary 2. Stability of predictions from approximate means.}
Fix the query, keys, and value vectors, with \(\|v_j\|\leq B\). Let \(p_J=(p_{J,j})_{j\in\mathcal A_i}\) be the exact source-probability vector and \(\widehat p_J\) an approximation, where
\[
p_{J,j}=P(J_i=j\mid q_i,K),\qquad
m=\sum_jp_{J,j}v_j,\qquad
\widehat m=\sum_j\widehat p_{J,j}v_j.
\]
The subscript \(J\) distinguishes source probabilities from output-token probabilities; \(j\) identifies one component. Under the exact Bayesian correspondence, attention weights satisfy \(\alpha_{ij}=p_{J,j}\); approximate attention weights play the role of \(\widehat p_{J,j}\).

Suppose the fixed prediction head \(G\) is \(L_G\)-Lipschitz; that is, there exists a finite constant \(L_G\geq0\) such that, for all input representations \(u\) and \(v\) in the domain under consideration,
\[
\|G(u)-G(v)\|\leq L_G\|u-v\|.
\]
This bounds its sensitivity: an input change cannot be amplified by more than \(L_G\) in the output. For example, \(L_G=2\) turns an input difference of \(0.01\) into an output difference of at most \(0.02\). Writing \(p_{\mathrm{out}}=G(m)\) and \(\widehat p_{\mathrm{out}}=G(\widehat m)\), we obtain
\[
\begin{aligned}
\|\widehat p_{\mathrm{out}}-p_{\mathrm{out}}\|
&\leq L_G\|\widehat m-m\|\\
&\leq2L_GB\,D_{\mathrm{TV}}(p_J,\widehat p_J)\\
&\leq L_GB\sqrt{2D_{\mathrm{KL}}(p_J\|\widehat p_J)}.
\end{aligned}
\]
The result traces source-weight error through the representation to output perturbation, complementing the loss decomposition. For a Transformer, these results distinguish three requirements: the representation supplied to the prediction head must retain the relevant contextual information, the head must use it accurately, and errors in attention weights should not cause large output changes. The stability bound addresses the third requirement under bounded-value and Lipschitz assumptions. 

\subsection{The information bottleneck connection}

The information bottleneck principle of Tishby et al.~\cite{tishby1999} asks how much of an input can be discarded while retaining information useful for prediction. Predictive sufficiency describes the ideal preservation condition: once the summary is known, seeing the full context would not improve the next-token distribution. Mutual information measures how much knowing one random variable helps predict another. For example, \(I(S_t;X_{t+1})\) measures how informative the summary is about the next token. It is symmetric, and is zero exactly when the variables are independent. For \(S_t=\Phi(X_{1:t})\), an idealized bottleneck objective is
\[ \min_{\Phi} \left[ I(S_t;X_{1:t}) - \beta I(S_t;X_{t+1}) \right]. \tag{23}\label{eq:paper-88} \]
The term ``bottleneck'' describes the summary as a restricted passage through which context information must reach the predictor:
\[
X_{1:t}\longrightarrow S_t\longrightarrow\text{prediction}.
\]
The predictor receives the context only through \(S_t\). The objective encourages this passage to retain less context information while preserving information useful for the next token. 

The minimization in Equation~\eqref{eq:paper-88} chooses a representation function \(\Phi\) that balances compression against predictive usefulness. The first term, \(I(S_t;X_{1:t})\), measures how much knowing the summary tells us about the original context \(X_{1:t}=(X_1,\ldots,X_t)\). If every context produces the same summary, it reveals nothing about which context occurred and this mutual information is zero. If the summary uniquely identifies the context, it retains all its information; otherwise, contexts sharing a summary are no longer distinguishable from that summary alone. For example, the count summary in Section 7.2 maps \(AAB\), \(ABA\), and \(BAA\) to the same value, discarding their order. Minimizing this term encourages removing such distinctions when they are unnecessary for prediction. The second term, \(I(S_t;X_{t+1})\), measures how much knowing the summary helps predict the next token compared with not knowing the summary. Without it, the ideal predictor uses the overall probabilities \(P(X_{t+1})\); with it, the predictor uses \(P(X_{t+1}\mid S_t)\). The expected improvement is
\[
I(S_t;X_{t+1})
=\mathbb E_{S_t}\!\left[
D_{\mathrm{KL}}\!\left(P(X_{t+1}\mid S_t)\,\middle\|\,P(X_{t+1})\right)
\right].
\]
This is the expected reduction in log loss obtained by knowing the summary. Because this term is subtracted in Equation~\eqref{eq:paper-88}, minimization rewards preserving next-token information. The parameter \(\beta\geq0\) sets the relative importance of the two terms. A small \(\beta\) favors compression, potentially sacrificing predictive information; a large \(\beta\) places more emphasis on next-token information, potentially retaining more of the context. 

We recall the chain rule for mutual information: for random variables \(A,B,C\),
\[
I(A;B,C)=I(A;B)+I(A;C\mid B).
\]
The information supplied by \(B\) and \(C\) together about \(A\) equals the information supplied by \(B\), plus the additional information supplied by \(C\) once \(B\) is known. Reversing the order gives
\[
I(A;B,C)=I(A;C)+I(A;B\mid C).
\]
The total is unchanged, although its division between the two terms may differ.

Applied to our representation, this rule separates the full context's predictive information into information retained by the summary and information lost when using that summary. Because \(S_t=\Phi(X_{1:t})\) is determined by the context, adding it to the full context supplies no new information:
\[
I(X_{t+1};X_{1:t},S_t)=I(X_{t+1};X_{1:t}).
\]
The chain rule expresses the same quantity by first revealing the summary and then revealing the full context:
\[
I(X_{t+1};S_t,X_{1:t})
=I(X_{t+1};S_t)+I(X_{t+1};X_{1:t}\mid S_t).
\]
Combining these identities gives
\[
\underbrace{I(X_{t+1};X_{1:t})}_{\text{full-context predictive information}}
=
\underbrace{I(X_{t+1};S_t)}_{\text{retained predictive information}}
+
\underbrace{I(X_{t+1};X_{1:t}\mid S_t)}_{\text{representation loss}}.
\]
The last term asks: after seeing the summary, how much more could we learn about the next token by seeing the full context? Any additional predictive information available from the context was lost when the context was replaced by the summary. As Section 7.4 establishes,
\[
I(X_{t+1};X_{1:t}\mid S_t)
=\mathbb E\!\left[D_{\mathrm{KL}}(p_t^*\|p_t^S)\right].
\]
Thus, representation loss is the expected additional log loss incurred by the best summary-based predictor compared with the best full-context predictor. This is the same representation-loss term that appears in Equation~\eqref{eq:paper-79}. Section 7.4 decomposes excess log loss into representation loss and prediction-head error; here, the chain rule decomposes full-context predictive information into retained information and representation loss. No head-error term appears in this information decomposition because it concerns the representation itself, independently of the trained head. The common term connects the two perspectives: predictive information discarded by the representation becomes unavoidable additional expected log loss, even for the best possible head. With the data distribution fixed, retaining more next-token information reduces this loss. It is zero exactly when the summary is predictively sufficient, so that \(I(S_t;X_{t+1})=I(X_{1:t};X_{t+1})\). The bottleneck objective balances compression against this preservation requirement.

In Section 7.2, the count difference \(N_A-N_B\) replaces the entire sequence without changing the next-symbol distribution. This illustrates compression that remains predictively sufficient under the specified generating model. For a Transformer, the corresponding question is whether its attention-based representation discards unnecessary details while retaining what determines the next-token distribution. 

\subsection{Algorithmic future-loss bounds and representation loss}

Solomonoff induction and its associated future-loss bounds offer a conditional route to bounding the next-token information lost by an attention representation. We specialize the continuation bound of Chernov et al.~\cite{chernov2007} to a single next token (\(m=1\)), combine it with an additional predictor-approximation assumption, and apply the decomposition in Section 7.4. 

Fix a context length \(n\). Let the sequence follow a computable probability measure \(\mu\) on a finite alphabet \(\mathcal X\), and let \(S_n=\Phi(X_{1:n})\) be a deterministic attention-based representation with learned parameters fixed. A normalized predictor \(r(a\mid S_n)\) assigns probabilities to the next token \(a\in\mathcal X\), using the context only through this representation. All logarithms and information quantities in this subsection use base 2.

Here, uppercase \(X_i\) denotes the random token at position \(i\), while lowercase \(x_i\) denotes a particular observed value. Thus \(X_{1:n}=(X_1,\ldots,X_n)\) is the random context and \(x=x_{1:n}=(x_1,\ldots,x_n)\) is an observed context. The bound involving \(B(x)\) below applies to a fixed observed context; \(\mathbb E_{X_{1:n}}\) subsequently averages over possible contexts according to their probabilities under \(\mu\).

\paragraph{The full-context result for one token.}
For an observed context \(x=x_{1:n}\) with \(\mu(x)>0\), Theorem 1 of Chernov et al.~\cite{chernov2007}, applied to a one-token continuation \(a\), gives
\[
\log_2\frac{\mu(a\mid x)}{M(a\mid x)}
\leq K(\mu\mid x)+K(n)+c_0=:B(x)
\]
whenever \(\mu(a\mid x)>0\). Here \(K(\mu\mid x)\) describes a shortest self-delimiting procedure computing \(\mu\) given \(x\), \(K(n)\) describes the context length, and \(c_0\) is a reference-machine constant independent of \(x,a,\mu\). The Solomonoff next-token weight is \(M(a\mid x)=M(xa)/M(x)\). Averaging over the next token bounds the ideal full-context predictor's expected additional log loss by \(B(x)\). 

\paragraph{The additional approximation requirement.}
Suppose a nonnegative penalty \(\eta(x)\) satisfies
\[
r(a\mid\Phi(x))\geq2^{-\eta(x)}M(a\mid x)
\]
for every context of positive \(\mu\)-probability and every token with \(\mu(a\mid x)>0\). This limits the representation-based predictor's additional log loss relative to \(M\), uniformly over those tokens. For example, \(\eta(x)=1\) requires at least half the reference weight. The penalty concerns both the representation and its head, not representation loss alone. Its smallest admissible nonnegative value for fixed \(r\) is
\[
\eta_r(x)=\max\left\{0,\max_{a:\mu(a\mid x)>0}
\log_2\frac{M(a\mid x)}{r(a\mid\Phi(x))}\right\},
\]
with an infinite ratio when the denominator is zero. 

\paragraph{From next-token loss to representation loss.} The full-context result for one token and the additional approximation requirement above give
\[
\log_2\frac{\mu(a\mid x)}{M(a\mid x)}\leq B(x),
\qquad
\log_2\frac{M(a\mid x)}{r(a\mid\Phi(x))}\leq\eta(x).
\]
Splitting the true-to-predicted log ratio at \(M\) combines these bounds:
\[
\begin{aligned}
\log_2\frac{\mu(a\mid x)}{r(a\mid\Phi(x))}
&=\log_2\frac{\mu(a\mid x)}{M(a\mid x)}
+\log_2\frac{M(a\mid x)}{r(a\mid\Phi(x))}\\
&\leq B(x)+\eta(x).
\end{aligned}
\]

Next, we average over possible next tokens by multiplying by \(\mu(a\mid x)\) and summing:
\[
\begin{aligned}
&D_{\mathrm{KL}}\!\left(\mu(\cdot\mid x)\,\middle\|\,r(\cdot\mid\Phi(x))\right)\\
&\quad=\sum_{a:\mu(a\mid x)>0}\mu(a\mid x)
\log_2\frac{\mu(a\mid x)}{r(a\mid\Phi(x))}\\
&\quad\leq B(x)+\eta(x).
\end{aligned}
\]
The right-hand side is unchanged because it does not depend on the next token and the averaging weights sum to one. Then, we average over contexts generated by \(\mu\), with \(S_n=\Phi(X_{1:n})\):
\[
\begin{aligned}
&\mathbb E_{X_{1:n}}D_{\mathrm{KL}}\!\left(
\mu(\cdot\mid X_{1:n})\,\middle\|\,r(\cdot\mid S_n)\right)\\
&\quad\leq\mathbb E_{X_{1:n}}[B(X_{1:n})+\eta(X_{1:n})].
\end{aligned}
\]
This bounds the representation-based predictor's total expected excess log loss.

Finally, we apply Section 7.4's decomposition, now with base-2 logarithms:
\[
\begin{aligned}
&\mathbb E_{X_{1:n}}D_{\mathrm{KL}}\!\left(
\mu(\cdot\mid X_{1:n})\,\middle\|\,r(\cdot\mid S_n)\right)\\
&\quad=I_\mu(X_{n+1};X_{1:n}\mid S_n)\\
&\qquad+\mathbb E_{S_n}D_{\mathrm{KL}}\!\left(
\mu(\cdot\mid S_n)\,\middle\|\,r(\cdot\mid S_n)\right).
\end{aligned}
\]
The first term is representation loss; the second is prediction-head error. Since the head-error term is nonnegative, representation loss alone cannot exceed their sum. Hence,
\[
I_\mu(X_{n+1};X_{1:n}\mid S_n)
\leq\mathbb E_{X_{1:n}}[B(X_{1:n})+\eta(X_{1:n})].
\]
A prediction head cannot recover information absent from its input. If two contexts require different next-token distributions but produce the same summary, a predictor using only that summary cannot distinguish them and must use the same prediction for both. This creates unavoidable additional expected log loss, even for the best possible head. Since representation loss is a nonnegative component of total expected excess log loss, a small total implies small representation loss. 

Here, the approximation assumption yields an upper bound of \(\mathbb E[B(X_{1:n})+\eta(X_{1:n})]\) on that total. The term \(B(x)\) bounds the reference predictor's additional loss, while \(\eta(x)\) bounds the summary-based predictor's further loss relative to that reference. The calculation provides a route to a guarantee only when the combined bound is sufficiently small; establishing such control for a trained Transformer remains an additional challenge. 

\section{Implications for Transformer Contextual Prediction}

We can now return to the motivating question: Why might attention be so effective for contextual prediction?

Our analysis suggests several possible contributions of attention to contextual prediction.

\paragraph{Attention evaluates relationships, not only token identities.}

A fixed embedding represents a token through learned numerical features. Attention makes the contribution of one token depend on its relationship to another token. The same word can therefore contribute differently to different predictions. This is important because language is relational. The meaning and predictive relevance of a token often depend on grammatical roles, references, surrounding words, and broader semantic structure. A mechanism capable of dynamically evaluating relationships is naturally suited to such tasks.

\paragraph{Attention implements soft selection.}

Attention does not generally require selecting exactly one source. It forms a weighted combination of candidates. Under the generative latent-source assumptions, this is posterior-weighted averaging of conditional payload means. Soft aggregation may be useful when several interpretations remain plausible or when information is distributed across multiple positions.

\paragraph{Attention supports conditional information aggregation.}

The Value vector contains information associated with a candidate source. The Query--Key operation determines the relative contribution of that information to a given query. Thus, attention separates two computational questions:
\begin{enumerate}
\item Which contextual information should contribute?
\item What information should be contributed?
\end{enumerate}
In our probabilistic model, these correspond to posterior source probabilities and conditional payload expectations. This separation provides a mechanism for input-dependent information aggregation.

\paragraph{Multi-head attention supports simultaneous relationships.}

Different attention heads can operate in different learned representation spaces. This enables several contextual relationships to be evaluated concurrently. Their outputs can then be combined and transformed. This permits complementary contextual computations.

\paragraph{Depth reorganizes information for prediction.}

With model parameters fixed, suppose one layer transforms \(H^{(\ell)}\) into \(H^{(\ell+1)}\). Here, \(H^{(\ell)}\) includes the representations at all token positions, and the layer receives no additional input beyond this state and fixed architectural quantities. Since the later representation is computed from the earlier one, it cannot contain information about the next token that was absent from the entire earlier representation. Formally, the data-processing inequality~\cite{coverThomas2006} gives
\[
I(X_{t+1};H^{(\ell+1)})
\leq I(X_{t+1};H^{(\ell)}).
\]
Later layers can nevertheless combine and reorganize existing information so that the available prediction head can use it more effectively. Attention may bring information from several positions into the representation used for prediction, and subsequent transformations can make relevant distinctions easier to extract. Information at an individual position may therefore increase even though its total amount across the complete representation cannot. Depth can improve the accessibility of predictive information without increasing its total amount.

\section{Limitations and Open Questions}

The main limitation of this analysis is the gap between computations that attention can implement under a specified probabilistic model and those acquired by a trained Transformer. Bridging this gap requires theoretical analysis and empirical evidence about the learned representations; their roles cannot be inferred uniquely from numerical weights or predictive behavior alone. The categorical-source and Gaussian constructions provide tractable models, but their assumptions need not describe trained heads. Although a source’s value may already combine information from several positions, a single source variable does not necessarily capture all relationships involved in prediction. Even when attention computes an exact posterior mean, this does not establish predictive sufficiency or accurate use of the representation by the prediction head. Likewise, additional layers can reorganize information without guaranteeing improvement or automatically implementing further Bayesian updates. The next task is therefore to identify assumptions about the architecture, data distribution, and training procedure under which useful representations and accurate heads can be learned from finite samples. A complete account must connect the data distribution and learning procedure to the information retained by the representation and the accuracy of its prediction head. Any resulting guarantee should specify the attention architecture, data-generating model, training procedure, and reference predictor. The algorithmic prediction connection additionally requires a useful bound on the approximation penalty, which remains open for trained attention representations. The stability bounds in Sections~\ref{subsec:source-stability} and~\ref{par:prediction-stability} control perturbations but do not establish these learning or predictive guarantees. Establishing conditions that connect information preservation, prediction accuracy, learnability, and efficiency remains the central research challenge.

\section{Conclusion}

This paper connects existing probabilistic interpretations of Transformer attention with predictive sufficiency and representation loss. Under explicit assumptions, attention’s weighted value aggregation can be interpreted as a conditional expectation and, in a generative model, as a Bayesian posterior mean. Building on Gabbur et al.’s framework, the Gaussian construction recovers scaled dot-product attention for arbitrary key norms. The central implication is that optimal estimation of a contextual payload does not alone establish optimal next-token prediction: the representation must preserve the relevant information, and the prediction head must use it correctly. An established excess-log-loss decomposition makes this distinction precise by separating representation information loss from prediction-head error, while stability bounds control the effects of source-weight errors under bounded-value and Lipschitz assumptions. The information bottleneck perspective relates compression to predictive preservation, and the connection to Solomonoff prediction identifies an additional approximation requirement. Together, these analyses help identify the assumptions under which probabilistic interpretations can support useful guarantees for Transformer attention. Determining which conditions hold in trained models, and how architecture and training might make them hold more reliably, could guide the development of AI systems that preserve and use predictive information more effectively.

\end{document}